\documentclass[10pt,letterpaper]{article}
\usepackage[top=0.85in,left=2.75in,footskip=0.75in]{geometry}

\usepackage{amsmath,amssymb}

\usepackage{changepage}

\usepackage[utf8x]{inputenc}

\usepackage{textcomp,marvosym}

\usepackage{cite}

\usepackage{nameref,hyperref}

\usepackage[right]{lineno}

\usepackage{microtype}
\DisableLigatures[f]{encoding = *, family = * }

\usepackage[table]{xcolor}

\usepackage{array}

\newcolumntype{+}{!{\vrule width 2pt}}

\newlength\savedwidth

\raggedright
\usepackage[aboveskip=1pt,labelfont=bf,labelsep=period,justification=raggedright,singlelinecheck=off]{caption}

\makeatletter
\renewcommand{\@biblabel}[1]{\quad#1.}
\makeatother

\usepackage{lastpage,fancyhdr,graphicx}
\usepackage{epstopdf}

\usepackage{adjustbox}
\usepackage{multirow}
\usepackage{makecell}
\usepackage{subfig}

\fancyheadoffset[L]{2.25in}
\fancyfootoffset[L]{2.25in}
\begin{document}
\vspace*{0.2in}

\begin{flushleft}
{\Large
\textbf\newline{Colonoscopy Polyp Detection and Classification: Dataset Creation and Comparative Evaluations} 
}
\newline
\\
Kaidong Li\textsuperscript{1},
Mohammad I. Fathan\textsuperscript{1},
Krushi Patel\textsuperscript{1},
Tianxiao Zhang\textsuperscript{1},
Cuncong Zhong\textsuperscript{1},
Ajay Bansal\textsuperscript{2},
Amit Rastogi\textsuperscript{2},
Jean S. Wang\textsuperscript{3},
Guanghui Wang\textsuperscript{4*}
\\
\bigskip
\textbf{1} Department of Electrical Engineering and Computer Science, The University of Kansas, Lawrence, KS, 66045 USA
\\
\textbf{2} Gastroenterology, Hepatology and Motility, The University of Kansas Medical Center, Kansas City, KS 66160
\\
\textbf{3} Department of Medicine, Washington University School of Medicine, Saint Louis, MO 63110
\\
\textbf{4} Department of Computer Science, Ryerson University, Toronto ON, Canada, M5B 2K3
\\
\bigskip

%
%





* Corresponding author. wangcs@ryerson.ca

\end{flushleft}
\section*{Abstract}
Colorectal cancer (CRC) is one of the most common types of cancer with a high mortality rate. Colonoscopy is the preferred procedure for CRC screening and has proven to be effective in reducing CRC mortality. Thus, a reliable computer-aided polyp detection and classification system can significantly increase the effectiveness of colonoscopy. In this paper, we create an endoscopic dataset collected from various sources and annotate the ground truth of polyp location and classification results with the help of experienced gastroenterologists. The dataset can serve as a benchmark platform to train and evaluate the machine learning models for polyp classification. We have also compared the performance of eight state-of-the-art deep learning-based object detection models. The results demonstrate that deep CNN models are promising in CRC screening. This work can serve as a baseline for future research in polyp detection and classification.


\section*{Motivation}\label{Intro}

Colorectal cancer (CRC) is one of the most common cancers diagnosed throughout the world \cite{thanikachalam2019colorectal, haggar2009colorectal}. From the data of both sexes combined, CRC contributes to \(10.2\%\) of all cancer cases in 2018 as the third most common cancer, following lung cancer (\(11.6\%\)) and breast cancer (\(11.6\%\)) \cite{bray2018global}. It is the second deadliest cancer in terms of mortality causing \(9.2\%\) of the total cancer deaths \cite{bray2018global}. According to the statistics \cite{haggar2009colorectal}, both male and female are almost affected equally. Nevertheless, despite the high incidence and mortality rates, the deaths caused by CRC have been decreasing with an accelerating decline rate since 1980 for both men and women \cite{acs2015factfig}. This trend mainly reflects the progress achieved in early detection and treatment. 

Early detection plays a significant role in fighting CRC. It not only brings down the mortality but also prevents excessive treatment cost by diagnosing before CRC spreads to distant organs \cite{simon2016colorectal}. According to \cite{haggar2009colorectal}, the stages at which the disease is diagnosed highly correlate to survival, with a \(90\%\) 5-year survival rate for the localized stage, \(70\%\) for the regional stage, and \(10\%\) for distant metastatic cancer. Another reason we should rely on early detection is due to the nature of the symptoms and development of CRC. Although no symptoms can be easily observed before the tumor reaches a certain size (typically several centimeters) \cite{simon2016colorectal}, it would typically take several years to as long as a decade for CRC to develop \cite{stracci2014colorectal}, starting from precancerous polyps. Both facts add up to show the significance and potential of diagnosing CRC by regular screening at an early stage,  even before polyps become cancerous.

\subsection*{CRC screening options}
There are several common CRC screening options, which can be roughly divided into two categories: visual examinations and stool-based tests. Each method has its advantages and limitations. The evaluation needs to take into account a broad range of factors including statistic data and psychological effects. The most important metric, like many other screening tests, is `{sensitivity}' \cite{simon2016colorectal}, which is also called `recall' in some other fields, determined as the percentage of patients with the disease that is actually detected. From sensitivity, we know the possibility of a patient walking out of the clinic with lesion undetected, the consequences of which are severe. Therefore, in many instances, it is the single most important metric to optimize.

Another statistical measure that often comes along with sensitivity is `{specificity}', which is measured as the fraction of healthy people that are correctly identified. It indicates the potential of a test to falsely detect lesions in healthy clients. This will cause mental stress on the clients, and the following treatment might result in unnecessary physical harm and financial burdens. Thus, a high specificity test is also preferred. For a screening method in real clinical settings, there is generally a trade-off between the sensitivity and the specificity. With the consequences of missing a lesion much more grave than false diagnosis, sensitivity is usually preferred over specificity. A higher specificity screening can always follow a high sensitivity test to filter out the falsely diagnosed cases \cite{simon2016colorectal}. Other factors include how easy the preparation is, how accessible the facility is, how much the test costs, etc. Since individuals need to be screened are oftentimes asymptomatic, the experience will affect their compliance, which is an important part of an effective screening program \cite{simon2016colorectal}. In the following session, some common CRC screen methods and their properties are discussed. 

\textbf{Colonoscopy} is the recommended CRC visual examination screening method. The advantages of colonoscopy include high sensitivity, ability to remove lesions at detection and full access to proximal and distal portions of the colon \cite{simon2016colorectal}. The colonoscopy can reach a sensitivity of \(95\%\) in detecting CRC according to Rex {\it et al.} \cite{rex1997relative}. The disadvantages are mostly related to the way colonoscopy is conducted \cite{regula2006colonoscopy, simon2016colorectal}. At least one day before the test, it requires a complicated bowel preparation, which requires the participant to change diet and take medicine to cause diarrhea. During the test, sedation or anesthesia might be performed, and there is a risk of post colonoscopy bleeding. Thus, the suggested 10-year screening interval has a low compliance rate \cite{simon2016colorectal}. Narrow-Band Imaging (NBI) is a newly developed technique by modifying light source using optical filters in an endoscope system \cite{machida2004narrow}. Compared to normal colonoscopy, intensified lights of a certain wavelength can better present the mucosal morphology and vascular pattern \cite{chang2009comparative}. Studies show that NBI perform better in CRC detection than conventional colonoscopy \cite{machida2004narrow, chang2009comparative}.

\textbf{Computed Tomography (CT) Colonoscopy} is a structural radiologic examination that employs software to reconstruct 3D views of the entire colon to detect lesions. Although it has a slightly less sensitivity of \(>90\%\), the less-invasive nature of CT colonoscopy results in a higher participation rate \cite{johnson2008accuracy, stoop2012participation}. The limitations include unpleasant bowel preparation before the test, uncomfortable inflation of colon with air during the test, and safety concerns over the use of radiation. Compared to colonoscopy, CT colonoscopy is not studied thoroughly, {\it e.g.}, uncertain screening interval \cite{simon2016colorectal}. With the fact that CT colonoscopy requiring follow-up colonoscopy with lesion detected and its sensitivity highly dependant on radiologists' expertise, it is only recommended to individuals whose physical conditions are not fit for the invasive examination of the colon \cite{simon2016colorectal}. A similar screening method, double-contrast Barium Enema, is also not recommended due to similar limitations and even more complicated procedures \cite{lieberman2009screening}.

\textbf{Sigmoidoscopy} is similar to colonoscopy, but it can only access the distal part of the colon. It shares the same high sensitivity as colonoscopy and can remove lesion at the detection. In addition, it requires less complicated bowel preparation and usually does not need sedation \cite{lieberman2009screening}. However, sigmoidoscopy has limited accessibility to only the distal colon rather than the proximal part, making it less effective due to the higher risk of proximal CRC among elder individuals and women \cite{lieberman2009screening}. Therefore, it is recommended to pair sigmoidoscopy with other screen methods \cite{simon2016colorectal}.

\textbf{Wireless Capsule Endoscopy} uses a miniaturized camera in a swallowable capsule to transmit gastrointestinal images to portable receiver units that can be easily worn \cite{iddan2000wireless}. Although a typical examination takes about 7 hours \cite{adler2003wireless}, the process does not impact patients' life quality compared to other methods. This wireless capsule can also examine the entire small bowel that is not accessible to other endoscopy practices \cite{adler2003wireless}. Nevertheless, wireless capsule endoscopy has some drawbacks as well. For example, it has no therapeutic capability \cite{adler2003wireless}. Also, it does not take images in distended bowel as other methods \cite{adler2003wireless}, the practitioners need training to interpret the images.

\textbf{Fecal Occult Blood Test (FOBT) and Fecal Immunochemical Test (FIT)} both detect hemoglobin in the stool to indicate if a lesion exists. Both tests are non-invasive and easy to be carried out even at home, but their sensitivities suffer for earlier stages of lesions due to less frequent bleeding \cite{simon2016colorectal}. In addition, some dietary intakes can alter the test results, reducing the performance of FOBT and FIT. 

There are other screening tests like the  DNA test, wireless capsule endoscopy, etc. However, due to low sensitivity and lack of sufficient supportive studies, they normally need a subsequent colonoscopy when the result is positive.

\subsection*{Goals}
As the reference CRC screening test, colonoscopy has obvious advantages over its alternatives. However, its performance depends on several variables, like the bowel preparation, the number of polyps, and the part of the colon where the polyps are located \cite{chan2009fewer, leufkens2012factors, patel2016real}. Furthermore, human-factors can influence the screening sensitivity and specificity. Inexperienced gastroenterologists have higher miss detection rates compared to those who are well-trained. According to Leufkens {\it et al.} \cite{leufkens2012factors}, participants before training showed significantly lower performance than post-training results. Colonoscopy is also subjected to the physical and mental fatigue of the gastroenterologists. The screening process requires prolonged concentration and is usually repeated throughout the day. A study by Chan {\it et al.} \cite{chan2009fewer} showed that \(20\%\) more polyps are detected from early morning screenings.  

It is obvious that a fine-grained deep learning framework to automatically detect polyps is needed to help physicians locate and classify the lesions. This deep learning framework can assist physicians during screening in real-time and prompt the detected region and polyp category. Thus, such a computer-aided system can help eliminate the miss rate due to physical and mental fatigue and allow the gastroenterologists to focus on regions where lesions actually exist. This automated system also ensures high performance in clinics where access to experienced gastroenterologists is difficult. An accurate detection system can also improve the detection rate of smaller pre-cancerous polyps using the Convolutional Neural Network (CNN) models. The sensitivity of current colonoscopy suffers as the size of the colon becomes smaller \cite{lieberman2009screening, stracci2014colorectal, simon2016colorectal}. This can be improved because the state-of-the-art CNN models can extract features from objects at different scales.

Deep learning models require larger datasets to exploit its full potential. Recent benchmark datasets for general computer vision tasks all have more than \(10k\) images \cite{Everingham10}. We want to build a polyp classification dataset based on the videos from the colonoscopy procedure with a reasonable number of samples to train deep neural network models. The images in the dataset contain polyps from different stages and are representative of different types of polyps. We will label each frame with accurate polyp locations and categories. Although constructing such a dataset is time-consuming and labor-intensive, it will benefit the research community to develop more accurate and robust deep learning models to achieve a higher detection rate and to reduce CRC mortality rate. The dataset could also standardize and facilitate the training of medical professionals in endoscopy.

Using the developed dataset, we have evaluated and compared the performance of the state-of-the-art deep learning models for polyp detection and classification. The dataset and the corresponding annotations can be downloaded via \href{https://doi.org/10.7910/DVN/FCBUOR}{\textit{https://doi.org/10.7910/DVN/FCBUOR}}.

\section*{Related Work}

Deep learning has achieved more and more attention in recent years with wide applications across a variety of areas. It boosts the performance by a significant margin in tasks like computer vision, speech recognition, natural language processing, data analysis, etc. \cite{li20202, krizhevsky2012imagenet, hinton2012deep, zhang2018bpgrad, najafabadi2015deep, li2021sgnet}. The success is largely owing to the development of deep Convolutional Neural Networks (CNN) which have been proven to be especially effective in extracting high-level features. Among all these areas, deep learning has achieved huge success in computer vision applications, with early CNN models almost halving error rate in the ImageNet classification challenge compared to classic models \cite{krizhevsky2012imagenet}. In recent years, CNN-based models have demonstrated their outstanding capabilities in many complicated vision tasks, like object detection, image segmentation, object tracking, etc. \cite{ma2020mdfn, huang2019mask, bertinetto2016fully, sajid2020plug, xu2019adversarially}.

\subsection*{Computer Vision in Medical Applications}
Researchers have been trying to use computer vision techniques in medical applications as early as 1970 \cite{litjens2017survey}. At that time, image processing was only a low-level task like edge finding and basic shape fitting. As the handcrafted models became more sophisticated, some studies showed success in areas like salient object detection and segmentation \cite{zhang2015minimum, huo2018supervoxel}. The ability of these models to analyze the surface pattern and appearance prompts their application in a wide range of medical fields, such as neuro, retinal, digital pathology, cardiac, and abdominal \cite{litjens2017survey}. Bernal \textit{et al.}  \cite{bernal2012towards} proposed a model that considers polyps as protruding surfaces and utilize valley information along with completeness, robustness against spurious responses, continuity, and concavity boundary constraints to generate energy map related to the likelihood of polyp presence. In the study \cite{karkanis2003computer}, the model exploits the color feature extraction scheme based on wavelet decomposition and then uses linear discriminant analysis to classify the region of interest. Other handcrafted feature approaches can be found in \cite{taha2017automatic}.

The limiting factor of hand-engineering models is the need for researchers to understand and design filters. They tend to perform better for low-level features. Deep learning models can automatically generate parameters with deeper layers and extract high-level semantic features. Especially in recent years, many new models \cite{cai2018cascade, zhao2019m2det} and techniques \cite{cen2021deep, wu2019unsupervised, zhang2020self, xu2020adaptively} have been published to set new records in various computer vision tasks. \cite{park2015polyp} employs multi-scale architecture with 3 layers of CNN and 3 layers of max-pooling followed by fully connected layers. Another model uses a slightly different approach using 3 different extracted features, color and texture clues, temporal features, and shape to feed an ensemble of 3 CNN models \cite{tajbakhsh2015automatic}. Deep learning models have been widely applied to medical problems like anatomical classification, lesion detection, and polyp detection and classification in colonoscopy \cite{patel2020comparative, roth2015anatomy, roth2015detection, pappalardo2020exploitation, mathew2020augmenting, mo2018efficient}. In \cite{patel2020comparative}, Six classical image classification models have been compared to determine the categories of detected polyps. It assumes all polyps have been detected and cropped out from the original sequences.
An enhanced U-Net structure has been proposed in \cite{patel2021enhanced} for polyp segmentation. In this paper, we focus on polyp detection from the endoscopic sequences to assist gastroenterologists in both polyp detection and classification. We evaluate and benchmark the state-of-the-art detection models for colonoscopy images.

\subsection*{Object Detection}
Different computer vision techniques can be adapted to perform polyp detection, such as object detection, segmentation, and tracking. {Object detection} takes images as input and generate classification results of objects presented in the images and their corresponding location information. Object locations are most commonly defined by rectangular bounding boxes. The output of image segmentation contains more detailed information, such as the classification result for each pixel in the original photo, while object detection usually only produces the coordinates of four corners of each bounding box. Thus, image segmentation is usually more time-consuming. In practice, pixel-level classification is not necessary for polyp detection and classification. In this study, we focus on object detection techniques. The state-of-the-art deep learning-based object detection models can be broadly classified into two main categories: two-stage detectors and one-stage detectors.

\textbf{Two-stage detector} consists of a region proposal stage, followed by a classification stage. Each of the two stages has its own dedicated deep CNN, which generally produces higher accuracy compared to one-stage detectors. However, this also leads to more processing time. The region proposal stage used to be the bottleneck as it is often a slow process, while the state-of-the-art two-stage detectors adopt new structures sharing part of the CNN to speed up the processing time for real-time applications \cite{ren2015faster}.

\textbf{One-stage detector} gets rid of the region proposal stage and fuses it with the classification stage, resulting in a one-stage framework. It directly predicts bounding boxes by densely sampling the entire image in a single network pass. With simpler architecture, it often achieves real-time performance. Although earlier models had lower detection accuracy than the two-stage detectors, they are catching up and now can produce comparable results.

\section*{Evaluation Models for Detection and Classification}
In this section, we will make a brief introduction of eight state-of-the-art object detection and classification models that are implemented and evaluated in this comparative study.

\textbf{Faster RCNN} \cite{ren2015faster}: Faster RCNN is a two-stage framework model and one of the families of RCNN networks \cite{girshick2014rich, girshick2015fast}. It improves the Fast RCNN network by replacing the slow selective search algorithm with a region proposal network, resulting in a faster detection rate. Furthermore, the region proposal network is trainable, which can potentially achieve better performance.

\begin{figure}
	\includegraphics[width=\columnwidth]{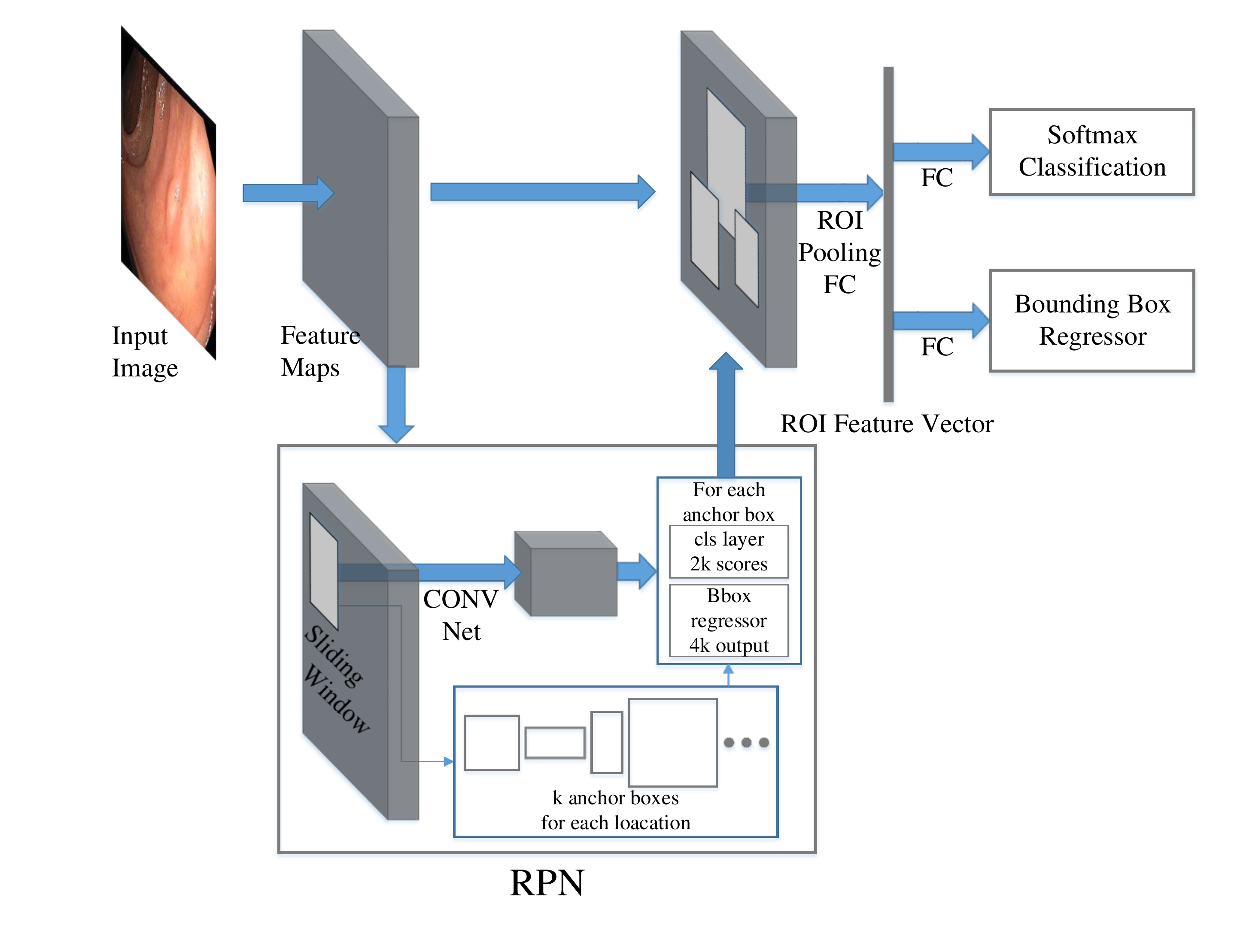}
	\centering
	\caption{\textbf{Faster R-CNN structure.} Region proposal network (RPN) shares the same base CNN with a fast R-CNN network. The region proposal is generated by sliding a small convolutional network over the shared feature maps, and these proposals are used to produce final detection results.}
	\label{fasterRcnnOverview}
\end{figure}

Faster RCNN is mainly composed of two modules, the region proposal network (RPN) module and the classification module, as shown in Figure~\ref{fasterRcnnOverview}. First, the backbone network (for example, ResNet 101 \cite{he2016deep}) extracts feature maps from the input image. The features are then shared by both the RPN module and the classification module. In the RPN branch, a sliding window will be applied to regress the bounding box locations and probability scores of object and non-object. At each location, the sliding window predicts $k$ pre-defined anchor boxes, centered at itself with different sizes and ratios to achieve multi-scale learning. With the introduction of RPN, the inference time on PASCAL VOC is reduced to $198 ms$ on a K40 GPU with VGG-16 as the backbone \cite{ren2015faster}. Compared to the selective search, it is almost $10$ times faster. The computational time of the proposal stage is reduced from $1,510 ms$ to only $10 ms$. Combined, the new faster R-CNN can achieve $5$ frames per second (fps).

\textbf{YOLOv3} \cite{redmon2018yolov3}: YOLOv3 is an iterative improvement of YOLO (You Only Look Once). It improves the performance of its previous versions by introducing a new backbone network, multi-scale prediction, and a modified class prediction loss function.

YOLO is the first model of this YOLO series \cite{redmon2016you, redmon2017yolo9000, redmon2018yolov3}. It is one of the pioneering works to get rid of the region proposal stage. The detector splits the image into $S \times S$ grids. Each cell is responsible for predicting ground truth objects with centers located inside the cell, and each cell in the grid predicts $ B \times (4 + 1 + C)$ values, where $B$ is the number of anchor boxes in each cell, $4+1$ represents the number of bounding boxes and object confidence, and $C$ is the total number of classes. The second version, YOLOv2, and YOLO9000, introduced several optimization tricks to improve the performance like batch normalization, high-resolution classifier, new network, multi-scale training, etc. Among the optimizations, the most effective technique is dimension priors which limit the regressed bounding boxes close to its original anchors. Without it, the regressed boxes can go anywhere in the image, resulting in unstable training \cite{redmon2017yolo9000}. YOLOv3 progressively developed a deeper CNN, DarkNet-53, from DarkNet-19 \cite{redmon2018yolov3}. It also predicts objects from different scales. YOLOv3 achieves real-time performance. However, it often has lower detection accuracy compared to Faster RCNN. 

\textbf{YOLOv4} \cite{bochkovskiy2020yolov4}: YOLOv4 is the latest improvement of YOLO. It explores the bag of freebies and bag of specials and selects some of them in the new detection model. The basic rules for a detection model are high-resolution input images for detecting relatively small objects, deeper layers for a larger receptive field, and more parameters for detecting various objects. Based on those rules, YOLOv4 selects various effective bag of freebies and bag of specials to enhance the performance of the model while maintains high-speed inference. In addition, instead of exploiting DarkNet53 as the backbone in YOLOv3, an enhanced version of DarkNet53 (CSPDarknet53 \cite{wang2020cspnet}) is selected as the backbone for YOLOv4. Higher receptive field is extremely important to detectors, thus SPP \cite{he2015spatial} net is added over the backbone CSPDarknet53 \cite{wang2020cspnet} since this block provides larger receptive fields with almost the same inference time. YOLOv3 utilizes FPN \cite{lin2017feature} to aggregate the information from different feature levels, while YOLOv4 \cite{bochkovskiy2020yolov4} employs PANet \cite{liu2018path} to extract information for detector heads. Bag of freebies and bag of specials are indispensable for object detection and properly selecting and adding them to the detection models may highly boost the performance of the detectors without sacrificing too much inference cost.

\textbf{SSD} \cite{liu2016ssd}: Single Shot Detector (SSD), as one of the most successful one-stage detectors, has become the foundation of many other studies. It takes advantage of the different sizes of feature maps and utilizes a simple architecture to generate predictions at different feature map scales. SSD can achieve a fast detection rate with competitive accuracy.

\begin{figure}
	\includegraphics[width=\columnwidth]{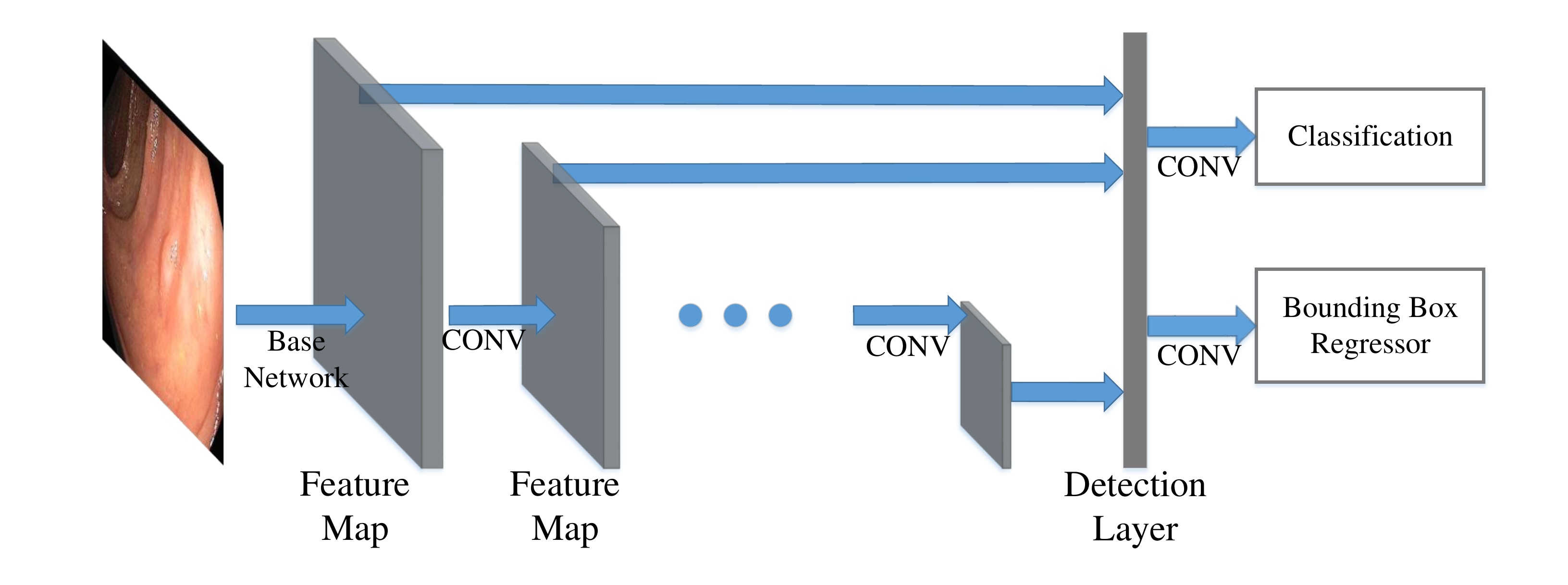}
	\centering
	\caption{\textbf{SSD structure.} Base network is truncated from a standard network. The detection layer computes confident scores for each class and offsets to default boxes.}
	\label{ssdArch}
\end{figure}

As shown in Figure~\ref{ssdArch}. SSD combines multi-scale convolutional features to improve prediction. In CNN, feature maps progressively decrease in size from input to output. The layers closer to input are shallow layers which have higher resolution and are better at detecting smaller objects. While the deeper layers have lower resolutions but contain more semantic information. SSD takes advantage of this natural structure of CNN and yields comparable results for objects with all sizes. SSD is an anchor-based detector. It divides the image into $m \times n$ grids similar to the YOLO series. At each grid cell, the model will generate per-class scores and bounding box dimension offsets for each $k$ pre-defined anchors with different ratios and scales, similar to RPN in Faster RCNN. It also introduces the use of convolutional layers for prediction which makes the detector fully convolutional, unlike YOLO \cite{redmon2016you} which uses fully connected layers for detection. 

SSD makes a good trade-off between speed and accuracy. The simple one-stage framework architecture results in fast performance, achieving a real-time detection rate. Furthermore, the use of anchor boxes and multi-scale prediction enables a good detection accuracy. 

\textbf{RetinaNet} \cite{lin2017focal}: RetinaNet is a one-stage framework based on the SSD model. RetinaNet improves performance by using the Feature Pyramid Network (FPN) \cite{lin2017feature} for feature extraction and focal loss function to solve the class imbalance problem. In the SSD model, the multi-scale prediction mechanism suffers from its architectural weakness in which high-level layers do not share information with low-level layers, thus lacking high-level semantic information in detecting smaller objects. FPN concatenate feature maps from layers at different depths to improve detection at each scale. Another major contribution of this model is the use of focal loss to solve the class imbalance problem. Class imbalance refers to the imbalance between background and foreground class. It is more extreme in one-stage models as the detector scans through the entire image indiscriminately. In practice, the candidate locations can normally go up to $100k$ without the filtering of the region proposal module. Therefore, the focal loss is introduced to assign higher weights to difficult foreground objects and lower weights to easy background cases. The definition of focal loss is defined in Equation~\ref{retNetFL}, where balance variant, $\alpha_t$, and focusing parameter, $\gamma$, are two hyper-parameters and $p$ is the estimated probability.

\begin{equation}
\label{retNetFL}
\begin{split}
FL(p_t)=&- \alpha_t (1-p_t)^\gamma log(p_t), \\
&\text{where }p_t = \left\{
  \begin{array}{ll}
    p, &\text{if correct detection}\\
    1-p, &\text{otherwise}\\
  \end{array}
\right.
\end{split}
\end{equation}

In Equation~\eqref{retNetFL}, $p_t$ is closer to $1$ when the model is more correct (i.e., correct prediction with higher confidence score or wrong prediction with lower confidence score). With the original cross entropy loss as $CE=- \alpha_t  log(p_t)$, focal loss effectively gives it a factor $(1-p_t)^\gamma$, whose value is small when the model is correct (easy cases) and large when the model is wrong (hard cases).

\textbf{DetNet} \cite{li2018detnet}: DetNet is a backbone network specifically designed to extract features, different from other detectors discussed in this section. It is designed to tackle three existing problems in previous backbone networks:

\begin{itemize}
\item Backbone networks have a different number of stages;
\item Feature maps used to detect large objects are usually from deeper layers, which have a larger receptive field, while they are not accurate in exacting the location due to low resolution;
\item Small objects are lost as the layers go deeper and resolutions become lower.
\end{itemize}

\begin{figure}
	\includegraphics[width=\columnwidth]{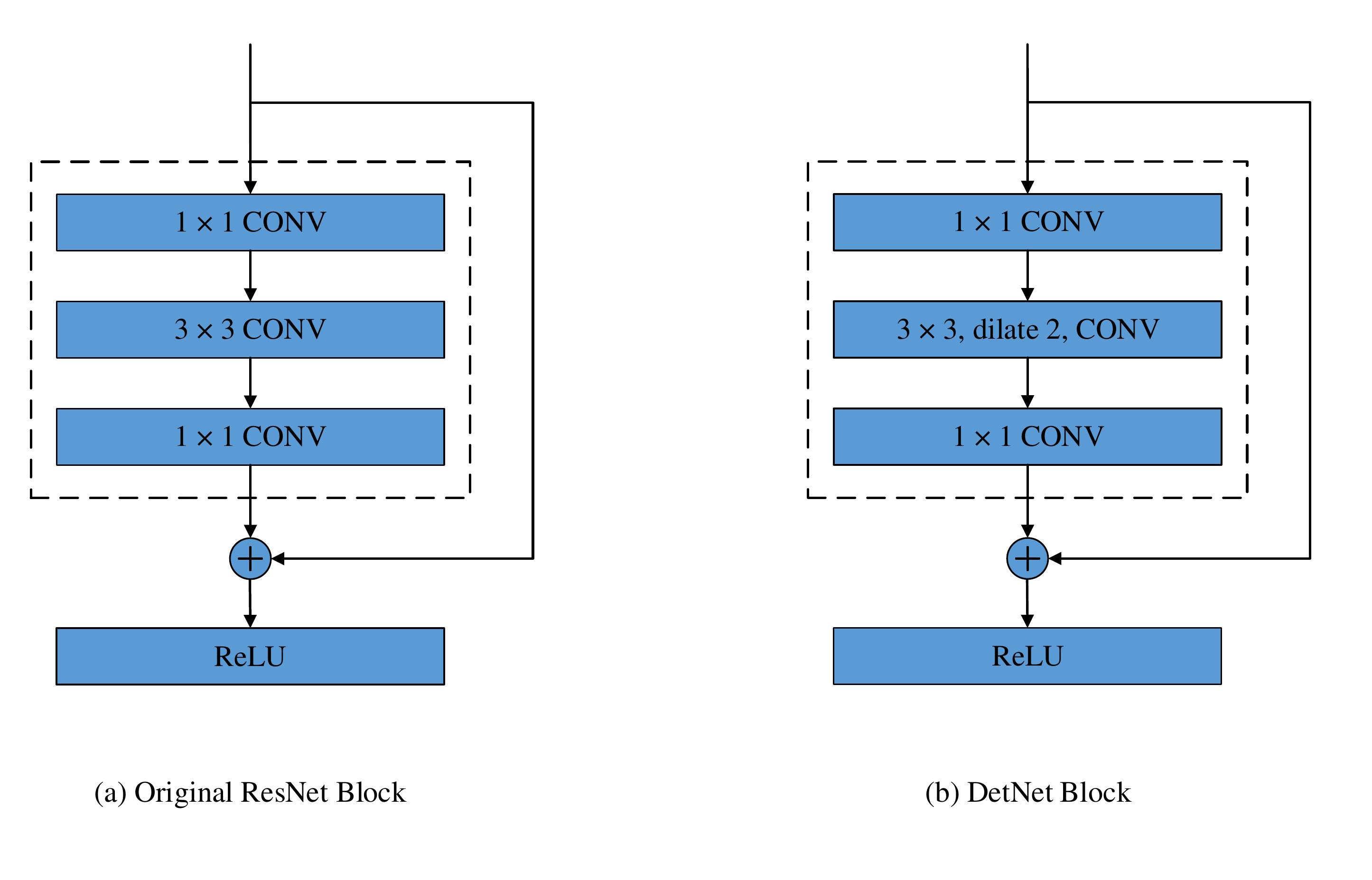}
	\centering
	\caption{\textbf{DetNet structure.} The diagram shows the basic building block of ResNet \cite{he2016deep} and DetNet \cite{li2018detnet}. (a) After each ResNet block, the resolution is reduce in half. (b) The DetNet preserves the feature map resolution and increases the receptive field by using dilated convolutions. }
	\label{detnet_block}
\end{figure}

Li {\it et al.} \cite{li2018detnet} proposed DetNet-59 based on ResNet-50. It has 6 stages with the first 4 stages the same as ResNet-50. In stages 5 and 6, the spatial resolutions are fixed instead of decreasing. The fixed resolution means a convolution filter will have a smaller receptive field compared to that in a lower resolution feature maps. A dilated \cite{yu2017dilated} bottleneck as shown in Figure~\ref{detnet_block}(b) is used for compensation. In this paper, we apply the DetNet backbone to the Faster RCNN detector.

\textbf{RefineDet} \cite{zhang2018single}: RefineDet is an SSD-based detector aiming at overcoming the following three limitations in single-stage detectors compared to the two-stage ones.

\begin{itemize}
\item Single-stage models lack region proposal module to eliminate the overwhelming background objects, causing inefficient learning;
\item Two-stage models have both region proposal module and classification module to regress final bounding box output while one-stage models only have one stage to refine box location;
\item Single-stage models generate only one set of feature maps for both tasks of localization and classification. Although recent two-stage models share the same backbone CNN, they have separate branches attached at the end of the main backbone networks for localization and classification specifically.
\end{itemize}

\begin{figure}
	\includegraphics[width=\columnwidth]{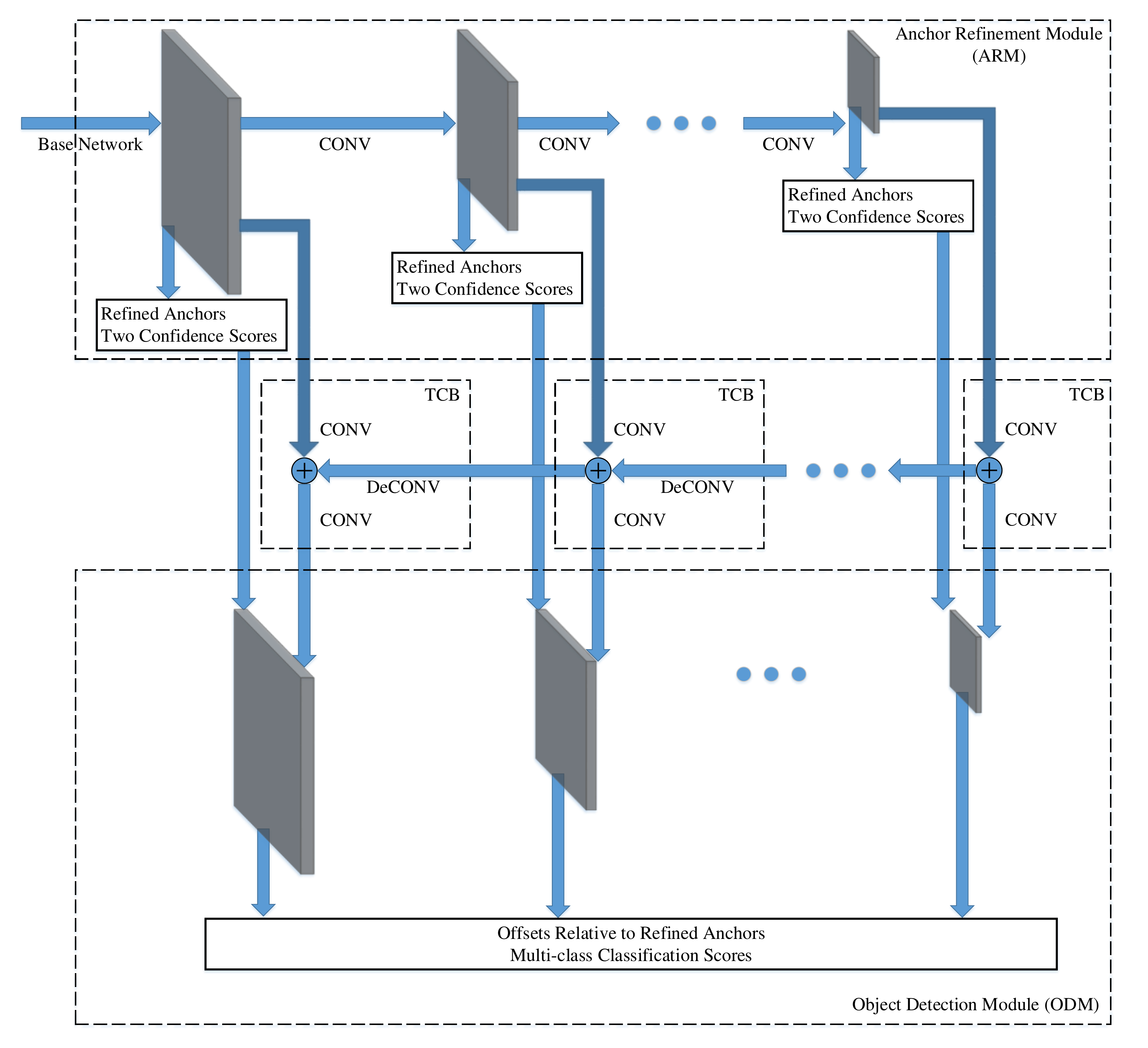}
	\centering
	\caption{\textbf{RefineDet structure.} The architecture has three modules: Anchor refinement module (ARM), transfer connection block (TCB), and object detection module (ODM).}
	\label{refine_chart}
\end{figure}

The architecture of RefineDet is shown in Figure~\ref{refine_chart}. It consists of three modules: Anchor refinement module (ARM), transfer connection block (TCB), and object detection module (ODM). Like in SSD, ARM takes feature maps from different layers. Then from each layer, it produces coarsely adjusted anchors and binary class scores (object and non-object classes). The anchors with a non-object score greater than a certain threshold $\theta$ will be filtered out, which reduces the class imbalance. Then the TCB is designed to combine features from deeper layers to the current level ARM features by element-wise addition. Deconvolution is used to facilitate the addition by increasing the resolution of deeper layer feature maps to match the shallow features. As a result, the shallow layers will have semantic information. By taking the filtered anchors from ARM and feature maps produced by TCB, ODM regresses the already refined anchors and generates multi-class scores. The results are improved because the input of ODM contains multi-level information and it refines the predicted bounding boxes in two steps.

\textbf{ATSS} \cite{zhang2020bridging}: ATSS (Adaptive Training Sample Selection) investigates the anchor-based object detectors and anchor-free object detectors and points out that how to define positive samples and negative samples in the training process is the significant difference between the anchor-based models and anchor-free models. For instance, the anchor-free detector FCOS \cite{tian2019fcos} first finds positive candidate samples in each feature level and then selects the final positive candidates among all features, while the anchor-based RetinaNet \cite{lin2017focal} exploits IoU (Intersection over Union) between pre-defined anchors and the ground truth bounding boxes to directly select the final positive samples among all feature levels \cite{zhang2020bridging}. Based on the analysis, ATSS automatically defines positive and negative candidates based on the statistical property of the objects in the images.

For each object on the image, ATSS selects $k$ anchor boxes based on the closest center distance between those samples and the ground truth box on each feature level. There are a total of $k\times L$ candidate positives if the number of feature pyramid levels is $L$. Then the IoU between these candidate samples and the ground-truth is calculated and the mean $m_g$ and standard deviation $v_g$ are also calculated so that the IoU threshold is obtained as $t_g = m_g + v_g$. Finally, the candidates whose IoU are larger than or equal to the threshold and at the same time whose centers are inside the ground-truth box are selected as the final positive samples. ATSS introduces a mechanism that dynamically selects the positive and negative samples and bridges the gap between anchor-based approaches and anchor-free approaches.

\section*{Dataset Build}

The performance of a CNN model is highly dependant on the dataset. During training, a CNN model learns from a large number of examples how to extract semantic features, on which localization and classification are based. Therefore, CNN detectors perform better when the dataset consists of representative examples of all categories. For example, images that are taken from different viewpoints, various illumination conditions, multiple sizes, etc. The more representative the dataset is, the more likely the CNN models can learn meaningful features for detection and classification. Then, at the inference time, the trained CNN models will have a higher ability to generalize the feature extraction on new input images. 

\begin{figure}
\centering
    \subfloat[MICCAI]{
        \includegraphics[height=2.6cm]{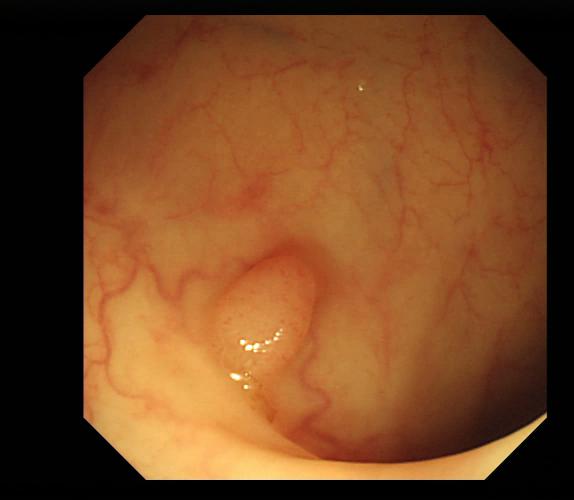} 
        \label{plots:psample_miccai}
    }
    \subfloat[GLRC]{
        \includegraphics[height=2.6cm]{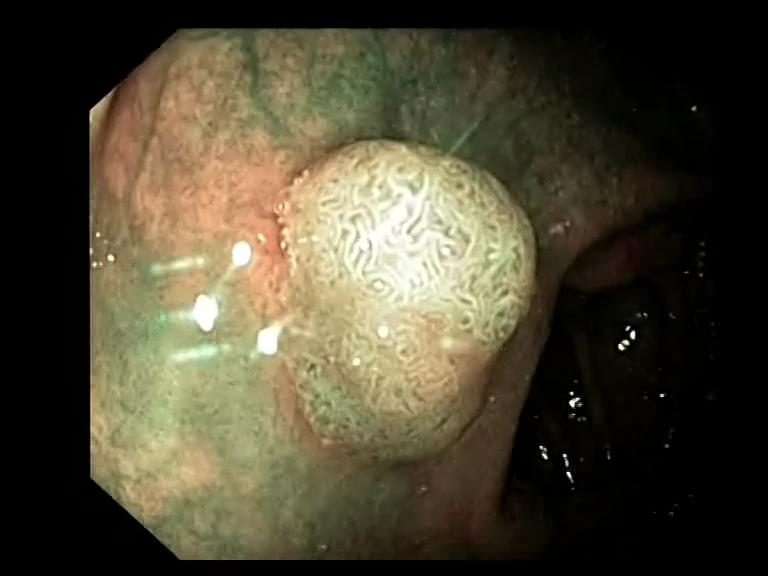} 
        \label{plots:psample_glrc}
    }
    \subfloat[KUMC]{
        \includegraphics[height=2.6cm]{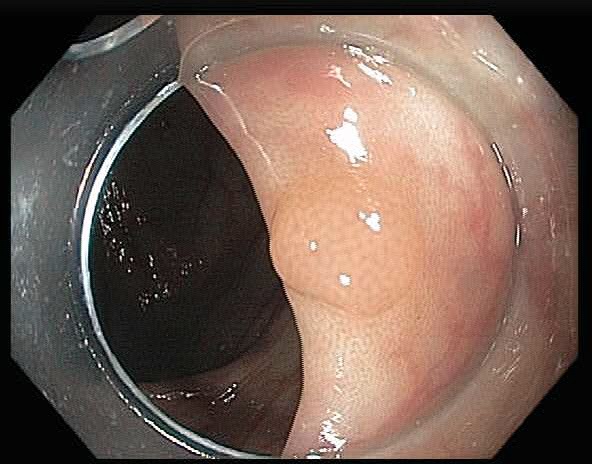} 
        \label{plots:psample_kumc}
    }
    
	\caption{Sample frames from different colonoscopy. (a) has a higher resolution and a warm color temperature; (b) has lower resolution and a green tone; (c) is more natural in color tone but has a transparent cover around the frame edges. }
	\label{plots:psample}
\end{figure}

In the research community, there are several small collections of endoscopic video datasets for different research purposes, such as MICCAI 2017, Gastrointestinal Lesions in Regular Colonoscopy Data Set (GLRC) \cite{mesejo2016computer} and CVC colon DB \cite{bernal2012towards} dataset. However, after careful observation and analysis, we found that these datasets differ greatly from each other in terms of resolution and color temperature, as shown in Figure~\ref{plots:psample}. This is largely due to the setups and characteristics of different imaging equipment used for data collection. As pointed out in \cite{chen2018domain}, two of the main reasons why current CNN models perform worse in the real-world compared to benchmark test sets are the variance in image backgrounds and image quality. As shown in Figure \ref{plots:psample}, the images in different datasets vary greatly. If we train the models using only one of these datasets, the models may have poor generalization ability, and their performance will suffer when being applied to colonoscopy images from different devices in another medical facility, as demonstrated in Section Experiments and Section Results and Analysis. More recently, there are several large dataset published on colonoscopy \cite{jha2021real, smedsrud2021kvasir}, like Hyper-Kvasir \cite{borgli2020hyperkvasir} and  Kvasir-SEG \cite{jha2020kvasir}. Hyper-Kvasir is a general-purpose dataset for gastrointestinal endoscopy. It detects 23 different classes of findings in the images and videos, including polyp, Angiectasia, Barretts, etc. \cite{borgli2020hyperkvasir} However, it does not provide the hyperplastic and adenomatous classification. Similarly, Kvasir-SEG provides labels in segmentation format. Thus, they could not be used to train detection models to predict polyp categories. 

\begin{figure}
\centering
    \subfloat[Frame 1]{
        \includegraphics[height=2.6cm]{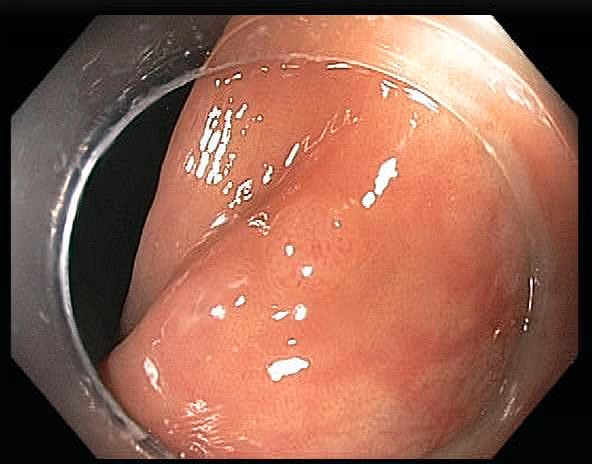} 
        \label{plots:frame_1}
    }
    \subfloat[Frame 70]{
        \includegraphics[height=2.6cm]{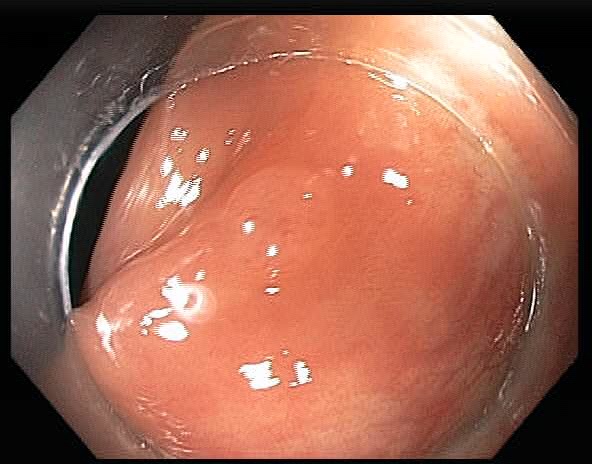} 
        \label{plots:frame_70}
    }
    \subfloat[Frame 146]{
        \includegraphics[height=2.6cm]{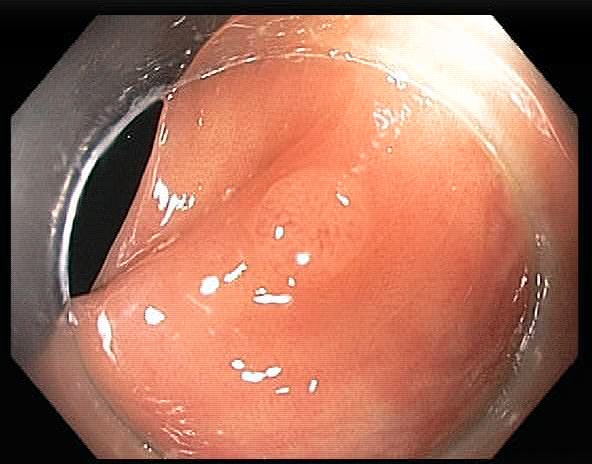} 
        \label{plots:frame_146}
    }
    
	\caption{A colonoscopy sequence. From frame 1 to frame 146, the camera shows unnoticeable movement.}
	\label{plots:frame_still}
\end{figure}

Another big limiting factor is the lack of distinct training examples. Although the available dataset seems to have many images, these images are actually extracted from a small number of video sequences. Each endoscopic video sequence only contains a single polyp viewed from different viewpoints. If we inspect the polyp frame by frame, we can see that most of the frames are taken from almost the identical viewpoint and distance as shown in Figure~\ref{plots:frame_still}. Some video sequences do not have noticeable movement across 1000 frames. Thus, there are significant redundancies in these datasets, especially for polyp classification, which required a large collection of distinct videos (polyps) to train the classifier. Considering recent benchmark datasets like MS COCO \cite{lin2014microsoft} with over 300k distinct images, more colonoscopy data are needed to achieve reasonable performance. 

In order to make the best use of the recent development of deep learning technologies for object detection. We collected and created an endoscopic dataset and compared the performance of the state-of-the-art detectors for polyp detection and classification. These datasets come from various sources and serve different purposes as will be discussed in the following subsection. To integrate them together, we refer to PASCAL VOC \cite{Everingham10} object detection task to standardize the annotation. The dataset only contains two categories of polyps: \textbf{hyperplastic} and \textbf{adenomatous polyps}. It is important to train a model that could reliably differentiate them since adenomatous polyps are commonly considered as precancerous lesions that require resection while hyperplastic polyps are not \cite{mesejo2016computer}.

\subsection*{Datasets Selection and Annotation}

In this study, we have collected all publicly available endoscopic datasets in the research community, as well as collected a new dataset from the University of Kansas Medical Center. All datasets are deidentified without revealing the patient information. With the help of three endoscopists, we annotated the polyp classes of all collected video sequences and the bounding boxes of the polyp in every frame. Below is an introduction to each dataset.

\textbf{MICCAI 2017:} This dataset is designed for Gastrointestinal Image ANAlysis (GIANA), a sub-challenge of the Endoscopic Vision Challenge \cite{bernal2017comparative}. It contains 18 videos for training and 20 videos for testing. The dataset is only labeled with polyp masks to test the ability to identify and localize polyps within images. There are no classification labels in this dataset. We converted the polyp masks into bounding boxes for each frame and annotated the polyp class.

\textbf{CVC colon DB:} The dataset has 15 short colonoscopy videos with a total of 300 frames \cite{bernal2012towards}. The labels are in the form of segmentation masks, and there are no classification labels. We extracted the bounding boxes and labeled the polyp class.


\textbf{GLRC Dataset: } The Gastrointestinal Lesions in Regular Colonoscopy Dataset (GLRC) contains 76 short video sequences with class labels \cite{mesejo2016computer}. There is no label for polyp location. We manually annotated the bounding box of each polyp frame by frame.

\textbf{KUMC Dataset: } The dataset was collected from the University of Kansas Medical Center. It contains 80 colonoscopy video sequences. We manually labeled the bounding boxes as well as the polyp classes for the entire dataset.

\subsection*{Frame Selection}
The video sequences from these datasets consist of different numbers of frames. For example, CVC colon DB only has 300 frames in total, averaging 20 frames per video sequence, while the number of frames in MICCAI 2017 varies from 400 to more than 1000 with a median value of around 300 in each sequence. The extreme imbalance among different lesions will reduce the representativeness of the dataset. In addition, many frames in a long sequence are redundant since they are taken with very small camera movement. To avoid some long videos overwhelming others, we adopt an adaptive sampling rate to extract the frames from each video sequence based on the camera movement and video lengths to reduce the redundancy and homogenize the representativeness of each polyp. After sampling, we extracted around 300 to 500 frames for long sequences to maintain a balance among different sequences, while for small sequences like CVC colon DB, we simply keep all image frames in the sequence.


\begin{figure}
\centering
    \subfloat[Underexposed]{
        \includegraphics[height=2.6cm]{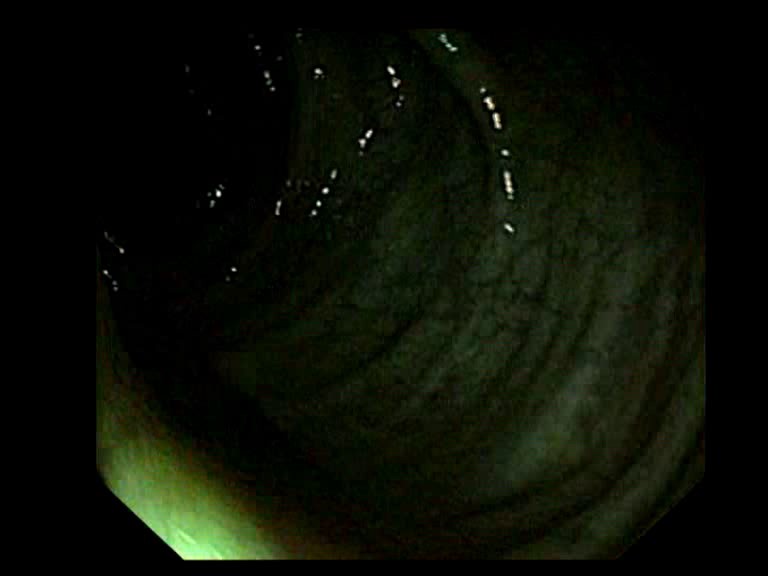} 
        \label{plots:bad_under}
    }
    \subfloat[Blurry]{
        \includegraphics[height=2.6cm]{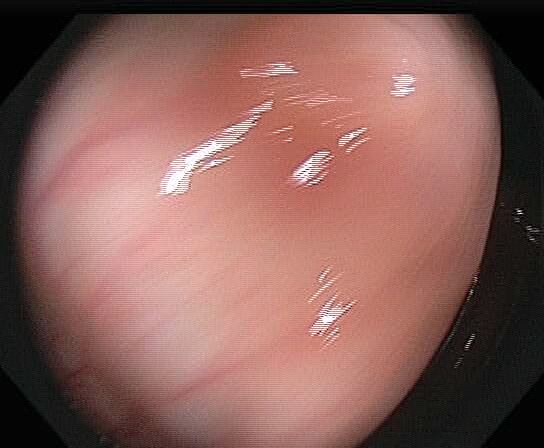} 
        \label{plots:bad_blur}
    }
    \subfloat[Out of Focus]{
        \includegraphics[height=2.6cm]{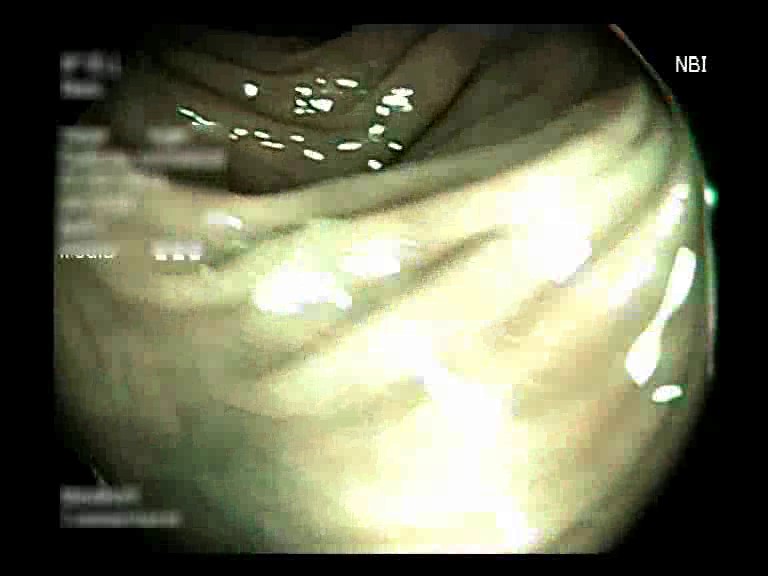} 
        \label{plots:out_focus}
    }
    
	\caption{Some bad examples of colonoscopy frames}
	\label{plots:frame_bad}
\end{figure}

After extracting all frames, we carefully checked the generated dataset and manually removed some frames that contain misleading or unuseful information. For example, when there is a sharp movement of the camera, the captured images may be severely blurred, out of focus, or subject to significant illumination change,  as shown in Figure~\ref{plots:frame_bad}. These images cannot be accurately labeled, so they are removed. While some less flawed frames are kept to improve the model's robustness under imperfect and noisy conditions. 

Polyp classification only by visual examination is a big challenge, as reported in \cite{mesejo2016computer}, the accuracy is normally below 70\% even for experienced endoscopists. In clinical practice, the results have to be confirmed by further biopsy tests. However, since we only have video sequences, when the endoscopist could not reach an agreement on the classification results, we simply remove those sequences from the dataset, otherwise, the models may not learn the correct information for classification. Eventually, the dataset contains 155 video sequences (37,899 image frames) with the labeled ground truth of the polyp classes and bounding boxes.

\subsection*{Dataset Split}


In order to train and evaluate the performance of different learning models, we need to divide the combined dataset into training, validation, and test sets. For most benchmark datasets for generic object detection, the split is normally based on images. However, this does not apply to the endoscopic dataset. Because all frames in one video sequence correspond to the same polyp, if we split the dataset at the image level, then the same polyp will simultaneously appear in the training, validation, and test sets. This will falsely increase the classification performance since the models have already seen the polyps to be tested during the training stage. Therefore, we split the dataset at the video level.

Since the final dataset is combined from four different datasets captured by different equipment with different data distribution. To increase the representativeness of the dataset, as well as the balance of the two classes of the polyps, we make the division for each dataset and polyp class independently. For each class in one dataset, we randomly select 75\%, 10\%, and 15\% sequences to form the training, validation, and test sets, respectively. For example, the GLRC \cite{mesejo2016computer} has 41 videos, with 26 adenomatous and 15 hyperplastic sequences. We split the 26 adenomatous sequences and the 15 hyperplastic sequences independently according to the same ratio to guarantee the class balance in the final dataset.

\begin{figure}
	\includegraphics[width=\columnwidth]{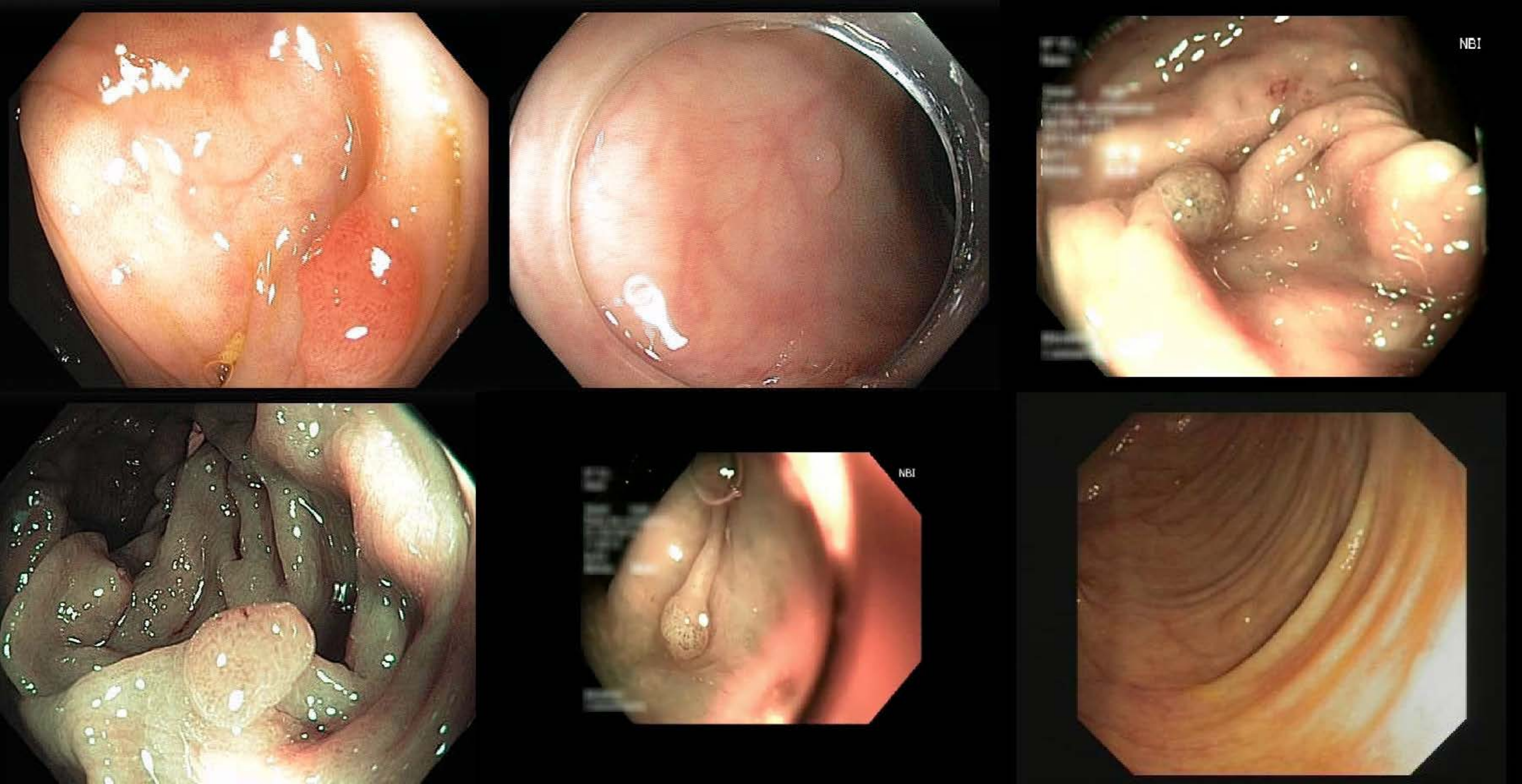}
	\centering
	\caption{\textbf{Six sample frames from the generated dataset.} }
	\label{plots:datasamples}
\end{figure}

In summary, we have generated 116 training, 17 validation, and 22 test sequences, with 28773, 4254, and 4872 frames, respectively, for each set. Some sample frames from the dataset are shown in Figure~\ref{plots:datasamples}. For the training set, we combine all frames from the 116 sequences into one folder and shuffle them. While for the validation and test sets, we keep the sequence split in order to evaluate the model performance based on polyps (i.e., sequences). The details of the dataset organization are shown in Table ~\ref{table:datasetBuild}. The dataset can be accessed from this \href{https://doi.org/10.7910/DVN/FCBUOR}{\textit{https://doi.org/10.7910/DVN/FCBUOR}}.

\begin{table*}
\begin{adjustwidth}{-2.25in}{0in}
\small
\centering
\caption{\textbf{Dataset Organization}}
\begin{adjustbox}{width=\textwidth+2.25in}
\begin{tabular}{ | c | r | r | r | r | r | r | r | r | r | r | r | r | r | r | r | r | r | r | r | r | r | r | r | r | }
    \hline
	Dataset   &\multicolumn{6}{c}{MICCAI2017}&\multicolumn{6}{|c}{CVC colon DB}&\multicolumn{6}{|c}{GLRC}& \multicolumn{6}{|c|}{KUMC 80} \\ 
	\hline
	\thead{before\\selection}  &    \multicolumn{6}{c}{38}    &     \multicolumn{6}{|c}{15}     & \multicolumn{6}{|c}{76} &   \multicolumn{6}{|c|}{80}    \\
	\hline
	\thead{after\\selection}  &    \multicolumn{6}{c}{23}    &     \multicolumn{6}{|c}{15}     & \multicolumn{6}{|c}{41} &   \multicolumn{6}{|c|}{76}    \\
	\hline
	\multirow{6}{*}{\# of Seqs} & \multicolumn{3}{c}{ad}  &  \multicolumn{3}{|c}{hp}  &  \multicolumn{3}{|c}{ad}  &  \multicolumn{3}{|c}{hp}  &  \multicolumn{3}{|c}{ad}  &  \multicolumn{3}{|c}{hp}  &  \multicolumn{3}{|c}{ad}  &  \multicolumn{3}{|c|}{hp}  \\
	\cline{2-25}
	&  \multicolumn{3}{c}{13}  &  \multicolumn{3}{|c}{10}  &  \multicolumn{3}{|c}{10}  &  \multicolumn{3}{|c}{5}  &  \multicolumn{3}{|c}{26}  &  \multicolumn{3}{|c}{15}  &  \multicolumn{3}{|c}{38}  &  \multicolumn{3}{|c|}{38}  \\
	\cline{2-25}
	&  \multicolumn{2}{c}{train}  &  \multicolumn{2}{|c}{val}  &  \multicolumn{2}{|c}{test} &  \multicolumn{2}{|c}{train}  &  \multicolumn{2}{|c}{val}  &  \multicolumn{2}{|c}{test}  &  \multicolumn{2}{|c}{train}  &  \multicolumn{2}{|c}{val}  &  \multicolumn{2}{|c}{test}  &  \multicolumn{2}{|c}{train}  &  \multicolumn{2}{|c}{val}  &  \multicolumn{2}{|c|}{test}  \\
	\cline{2-25}
	&  \multicolumn{2}{c}{20}  &  \multicolumn{2}{|c}{1}  &  \multicolumn{2}{|c}{2} &  \multicolumn{2}{|c}{11}  &  \multicolumn{2}{|c}{1}  &  \multicolumn{2}{|c}{3}  &  \multicolumn{2}{|c}{29}  &  \multicolumn{2}{|c}{5}  &  \multicolumn{2}{|c}{7}  &  \multicolumn{2}{|c}{56}  &  \multicolumn{2}{|c}{10}  &  \multicolumn{2}{|c|}{10}  \\
	\cline{2-25}
	&  ad  &  hp  &  ad  &  hp  &  ad  &  hp  &  ad  &  hp  &  ad  &  hp  &  ad  &  hp  &  ad  &  hp  &  ad  &  hp  &  ad  &  hp  &  ad  &  hp  &  ad  &  hp  &  ad  &  hp  \\
	\cline{2-25}
	&  12  &  8  &  0  &  1  &  1  &  1  &  7  &  4  &  1  &  0  &  2  &  1  &  19  &  10  &  3  &  2  &  4  &  3  &  27  &  29  &  5  &  5  &  6  &  4  \\
	\hline

\end{tabular}
\end{adjustbox}
\label{table:datasetBuild}
\end{adjustwidth}
\end{table*}

\section*{Experiments}\label{se:experiments}
Using the generated dataset, we evaluated eight state-of-the-art object detection models, including Faster RCNN \cite{ren2015faster}, YOLOv3 \cite{redmon2018yolov3}, SSD \cite{liu2016ssd}, RetinaNet \cite{lin2017focal}, DetNet \cite{li2018detnet}, RefineDet \cite{zhang2018single}, YOLOv4 \cite{bochkovskiy2020yolov4} and ATSS \cite{zhang2020bridging}. To set the benchmark performance, three different experiment setups are tested: \textbf{frame-based two-class polyps detection}, \textbf{frame-based one-class polyps detection}, and \textbf{sequence-based two-class polyps classification}. The performance of the two frame-based detections is measured using regular object detection metrics. For the sequence-based classification, regular detection models will be applied to each frame. Then a voting process picks the mostly predicted polyp category as the final classification result. More specific details will be presented below.

The eight detection models are mostly proposed with good performance on generic object detection tasks. These models are adopted from the originally published setups, with slightly modified hyperparameters to optimize their performance on the polyp dataset. The hyperparameter setups are listed in Table ~\ref{table:modelhyper}. We employ the following three metrics to evaluate the performance of each model: precision, recall, and F-score. 

\begin{table*}
\begin{adjustwidth}{-2.25in}{0in}
\small
\centering
\caption{\textbf{Experiment Setup}}
\begin{tabular}{ | c | r | r | r | r | r | r | r | r | }
    \hline
    
	&  Batch Size  &  Image Size  &  Learning Rate  &  Weight Decay  &  \thead{NMS\\Threshold}  &  \thead{Confidence\\Threshold}  &  \thead{Epoch/Iter\\(2-class det)} & \thead{Epoch/Iter\\(1-class det)}  \\
    \hline
	Faster RCNN  &    \(8\)  &  \(600\)  &  \(10^{-3}\)  &         \(10^{-1}\)  &  0.45  &  0.50  &   \(7\)         &   \(3\)        \\
    \hline
	SSD          &    \(8\)  &  \(300\)  &  \(4\times10^{-4}\)  &  \(10^{-4}\)  &  0.45  &  0.50  &   \(8\)         &   \(35k\) iter \\
    \hline
	YOLOv3       &   \(32\)  &  \(416\)  &  \(10^{-3}\)  &  \(5\times10^{-4}\)  &  0.45  &  0.50  &   \(24k\) iter  &   \(\)         \\
    \hline
	RetinaNet    &    \(1\)  &  \(608\)  &  \(10^{-5}\)  &                      &  0.50  &  0.50  &   \(9\)         &  \(1\)         \\
    \hline
	DetNet       &  \(8\)  &  \(600\)  &  \(10^{-3}\)  &         \(10^{-4}\)  &  0.45  &  0.50  &   \(2\)         &    \(5\)       \\
    \hline
	RefineDet    &    \(8\)  &  \(512\)  &  \(10^{-4}\)  &  \(5\times10^{-4}\)  &  0.45  &  0.50  &   \(35k\) iter  &   \(130k\) iter\\
    \hline
	YOLOv4    &    \(4\)  &  \(416\)  &  \(10^{-4}\)  &  \(5\times10^{-4}\)  &  0.45  &  0.50  &   \(40\)  &   \(26\) \\
    \hline
	ATSS    &    \(16\)  &  \(600\)  &  \(5\times10^{-3}\)  &  \(10^{-4}\)  &  0.45  &  0.50  &   \(15k\) iter  &   \(10k\) iter\\
    \hline

\end{tabular}
\label{table:modelhyper}
\end{adjustwidth}
\end{table*}

\begin{equation*}
\label{eq:metrics}
\begin{split}
Precision &= \frac{ \text{True Positive} }{\text{True Positive} + \text{False Positive}} \\
Recall &= \frac{ \text{True Positive}}{ \text{True Positive} + \text{False Negative}} \\
F_1 &= 2 \times \frac{Precision \times Recall}{ Precision + Recall }
\end{split}
\end{equation*}

\begin{itemize}

\item \textbf{Precision} measures the percentage of correct predictions. In polyp detection, it indicates the confidence in the prediction when a positive detection occurs. Higher precision can reduce the chances of a false alarm, which will cause the financial and mental stress of a client.
\item \textbf{Recall} is the fraction of the objects that are detected. It is very important in polyps detection since a higher recall ensures more patients receive a further check and appropriate treatment in time. It can also reduce mortality and prevent excessive cost to patients. 
\item \textbf{F-score} takes both precision and recall into consideration. It measures a balanced performance of a model between false positive and false negative.

\end{itemize}

\subsection*{Frame-based Two-class Polyp Detection}\label{sec.two-class} 
This experiment predicts polyps for individual frames. It is a test of a model's localization and classification ability. The CNN models are trained using our training set that consists of a mix of frames from different video sequences. During the validation and test phase, we treat each frame individually and evaluate the performance.

Since the state-of-the-art CNN detectors have fast detection speed and can be implemented in real-time. This allows the endoscopists to find the lesions and provide category suggestions during colonoscopy. As human operators may suffer from fatigue and focus loss after long hours of work, this automated process could alert and assist the endoscopists to focus on suspected lesions and avoid miss detection. 

To test the effectiveness of the proposed dataset with respect to a single dataset mentioned above, we also perform the frame-based two-class detection using a single dataset. In this controlled experiment, we train all the models trained using the KUMC dataset. Since this dataset contains a variety of more sequences and video frames than other datasets, it guarantees the convergence of all involved models. After training, we test the models on the same combined test set as in other experiments. As shown in the results, the performance of all models will drop significantly when trained using only a single dataset. This experiment verifies the effectiveness of the combined dataset.

\subsection*{Frame-based One-class Polyp Detection}\label{sec.one-class} 
This experiment has almost the same setups as the frame-based two-class polyp detection except for the class number. Hyperplastic and adenomatous polyps are treated as a single class \textit{polyps}. For annotation files, instead of providing a separate set of annotation files, we read the same ground truth as the previous experiment, discard the information about polyp categories during training and inference time. 

In colorectal cancer screening, it is more important to accurately detect whether polyps are developed than classifying polyp categories, because further screening and diagnosis are always followed after colonoscopy finds suspected lesions. This experiment aims to test whether a higher performance could be achieved by only localizing polyps in general. Without the more challenging task of classifying polyp categories, CNN models could be trained to extract more generalized features to distinguish polyps. Screening methods with higher precision like biopsy or polypectomy then could be followed to determine the categories of lesions.

\subsection*{Sequence-based Two-class Polyp Classification} 
This experiment adopts the same setup as Frame-based Two-class tests, however, we only make one prediction for each video sequence since it only contains the same polyp in the sequence. During the test period, it will generate the prediction based on individual frames at first, then we collect all results from every frame of a video sequence and classify the video based on the mostly predicted polyp category. Although there may have better ways to classify a video sequence such as based on the confidence score of the prediction for each frame, we only adopt the basic approach as a benchmark to see how much improvement we can achieve for sequence-based prediction.

Sequence-based classification is the practice of clinical application since all frames in the sequence are observing the same polyp from different viewpoints. It also has the potential to achieve better performance. To classify the polyp only from at frame is difficult, for example, the polyp may be partly occluded in some frames or appear small when viewing from a far distance. All these scenarios will make it hard to be accurately classified. However, in video-based classification, we are combining information from different viewpoints which can reduce the influence caused by those hard frames. Thus, at the clinic, the endoscopist usually takes the colonoscopic video from various viewpoints to ensure a reliable classification of the polyps.

\begin{figure}
\centering
    \subfloat{
        \includegraphics[height=3.5cm]{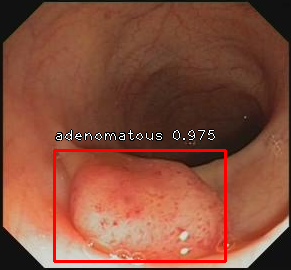} 
        \label{plots:vis1}
    }
    \subfloat{
        \includegraphics[height=3.5cm]{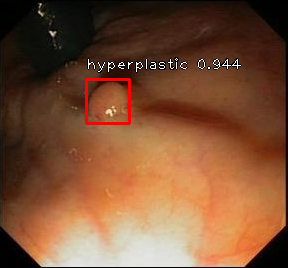} 
        \label{plots:vis2}
    }
    \subfloat{
        \includegraphics[height=3.5cm]{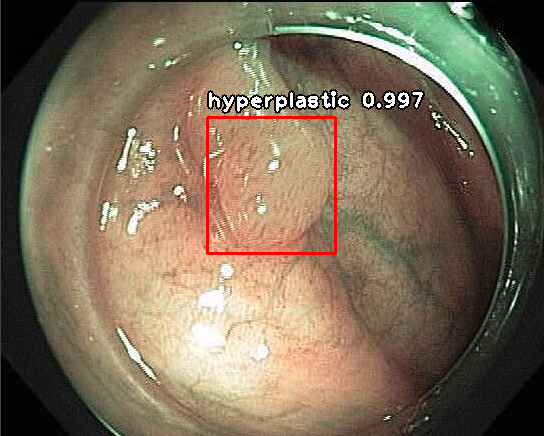} 
        \label{plots:vis3}
    }
	\caption{Three examples of the detection results with the predicted classes and confidence scores}
	\label{plots:example_res}
\end{figure}

\section*{Results and Analysis}\label{se:results}
In the experiments, the frame-based and sequence-based two-class detection and classification sharing the same CNN model. All hyperparameters for the compared models are summarized in Table~\ref{table:modelhyper}. The final models chosen for the test are based on the validation performance. Precision, recall, and F1 scores are all calculated at the confidence threshold of $0.5$ to ensure a fair comparison. The best performance CNN models are mostly produced before epoch $10$. An exception is RefineDet one-class detection, with $130k$ {\it iter} equaling around $45$ {\it epochs}. However, it has achieved similar validation performance, $88.05\%$ {\it mAP}, as early as $30k$ {\it iter} compared to $88.12\%$ at $130k$ {\it iter}. We suggest the best CNN model for polyps detection is usually generated at the earlier training stage.

\subsection*{Frame-based Two-class Polyp Detection} 
The results are shown in Table~\ref{table:resFTPD}. Overall, all detectors have achieved better performance for adenomatous polyps since they are larger in size and their shape and texture are easier to distinguish from the colonic wall. RefineDet has achieved the best combined performance. It achieves the highest mean F1-score, mAP, and mean recall than all other models. 
YOLOv3 yields the best precision by sacrificing its recall, which is abnormally lower than other detectors. Figure~\ref{plots:example_res} shows some examples of the detection results. We pick a confidence threshold of $0.5$. As shown in the examples, the models are very confident about the predictions. They mostly have only one prediction with a confidence score over $0.5$ on each frame. The predicted bounding boxes are very tight and precise on the lesions, which shows great potential in assisting colonoscopy practice. 

\begin{table}
\centering
\caption{\textbf{Results for Frame-based Two-class Polyp Detection}}
\begin{tabular}{ | c | c | r | r | r | r | }
    \hline
	                              & Category &  Precision &   Recall   &  F1-score &    AP   \\
    \hline \hline
	\multirow{3}{*}{Faster RCNN}  & ad &  \(72.8\)  &  \(73.0\)  &  \(72.9\)  &  \(72.9\)  \\
    \cline{2-6}
                                  & hp &  \(42.2\)  &  \(63.1\)  &  \(50.6\)  &  \(42.5\)  \\
    \cline{2-6}
                                  &Mean&  \(57.5\)  &  \(68.1\)  &  \(62.3\)  &  \(57.7\)  \\
    \hline \hline
    
	\multirow{3}{*}{SSD}          & ad &  \(82.7\)  &  \(77.4\)  &  \(80.0\)  &  \(82.7\)  \\
    \cline{2-6}
                                  & hp &  \(54.6\)  &  \(51.8\)  &  \(53.1\)  &  \(52.5\)  \\
    \cline{2-6}
                                  &Mean&  \(68.6\)  &  \(64.6\)  &  \(66.5\)  &  \(67.6\)  \\
    \hline \hline
    
	\multirow{3}{*}{YOLOv3}       & ad &  \(89.7\)  &  \(23.2\)  &  \(36.9\)  &  \(61.1\)  \\
    \cline{2-6}
                                  & hp &  \(60.0\)  &  \(16.2\)  &  \(25.5\)  &  \(35.0\)  \\
    \cline{2-6}
                                  &Mean&  \(\mathbf{74.9}\)  &  \(19.7\)  &  \(31.2\)  &  \(48.0\)  \\
    \hline \hline
    
	\multirow{3}{*}{RetinaNet}    & ad &  \(85.4\)  &  \(59.1\)  &  \(69.8\)  &  \(57.9\)  \\
    \cline{2-6}
                                  & hp &  \(52.9\)  &  \(43.7\)  &  \(47.9\)  &  \(40.5\)  \\
    \cline{2-6}
                                  &Mean&  \(69.2\)  &  \(51.4\)  &  \(59.0\)  &  \(49.2\)  \\
    \hline \hline
    
	\multirow{3}{*}{DetNet}       & ad &  \(73.0\)  &  \(67.5\)  &  \(70.2\)  &  \(60.4\)  \\
    \cline{2-6}
                                  & hp &  \(46.0\)  &  \(65.0\)  &  \(53.8\)  &  \(42.2\)  \\
    \cline{2-6}
                                  &Mean&  \(59.5\)  &  \(66.2\)  &  \(62.7\)  &  \(51.3\)  \\
    \hline \hline
    
	\multirow{3}{*}{RefineDet}    & ad &  \(92.2\)  &  \(61.3\)  &  \(73.6\)  &  \(81.1\)  \\
    \cline{2-6}
                                  & hp &  \(49.1\)  &  \(86.3\)  &  \(62.6\)  &  \(65.9\)  \\
    \cline{2-6}
                                  &Mean&  \(70.7\)  &  \(\mathbf{73.8}\)  &  \(\mathbf{72.2}\)  &  \(\mathbf{73.5}\)  \\
    \hline 

	\multirow{3}{*}{YOLOv4}    & ad &  \(90.5\)  &  \(54.0\)  &  \(67.6\)  &  \(70.4\)  \\
    \cline{2-6}
                                  & hp &  \(54.0\)  &  \(40.6\)  &  \(46.3\)  &  \(42.7\)  \\
    \cline{2-6}
                                  &Mean&  \(72.3\)  &  \(47.3\)  &  \(57.2\)  &  \(56.6\)  \\
    \hline 

	\multirow{3}{*}{ATSS}    & ad &  \(79.5\)  &  \(76.3\)  &  \(77.9\)  &  \(80.7\)  \\
    \cline{2-6}
                                  & hp &  \(57.2\)  &  \(68.0\)  &  \(62.2\)  &  \(58.4\)  \\
    \cline{2-6}
                                  &Mean&  \(68.4\)  &  \(72.2\)  &  \(70.2\)  &  \(69.5\)  \\
    \hline

\end{tabular}
\label{table:resFTPD}
\end{table}

\begin{figure}
    \subfloat[RefineDet]{
    \begin{centering}
        \includegraphics[width=\columnwidth]{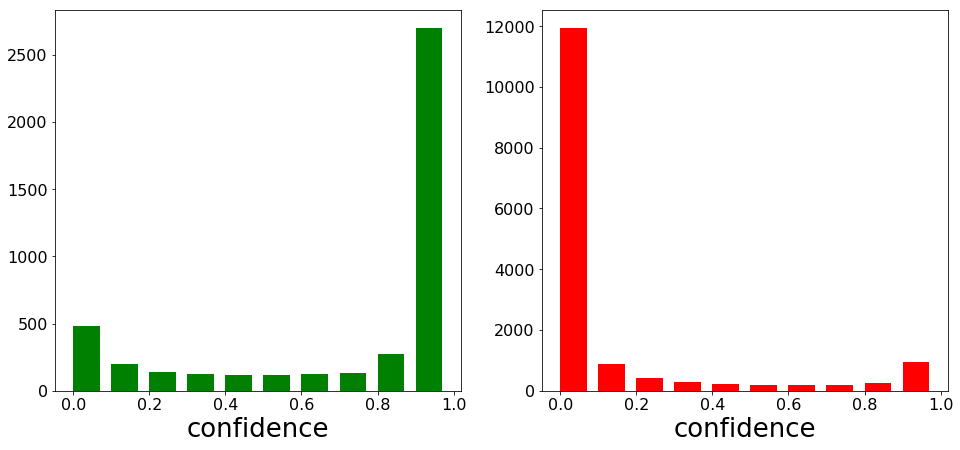}
        \label{plots:confidence_analysis_refine}
    \end{centering}}
    \hfil
    
    \subfloat[YOLOv3]{
    \begin{centering}
        \includegraphics[width=\columnwidth]{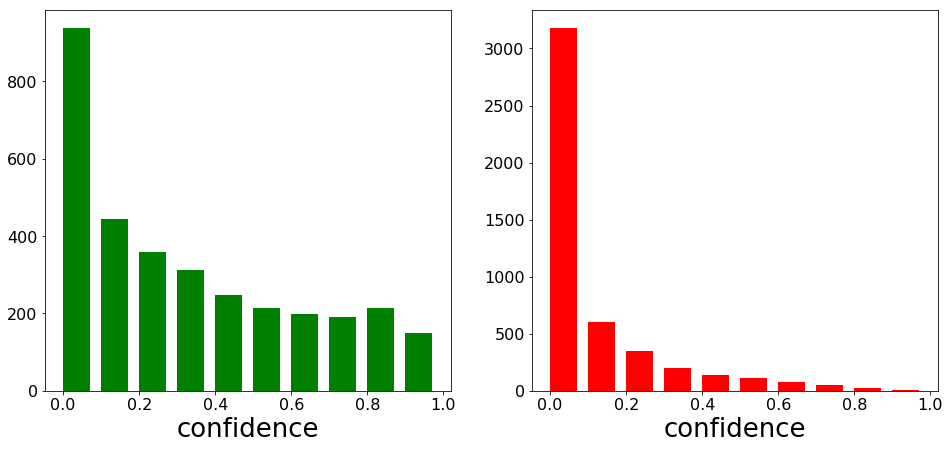} 
        \label{plots:confidence_analysis_yolo}
    \end{centering}}
    \hfil
    
	\caption{True positive (green plot) and false positive (red plot) count w.r.t. confidence. We discard any predictions with a confidence score below 0.01 since they tend to be random predictions.}
	\label{confidence_analysis}
\end{figure}

To analyze the difference between recalls from YOLOv3 and other detectors at the confidence threshold of $0.5$, we have plotted the count of true positives (TP) and false positives (FP) over different confidence scores. In Figure~\ref{confidence_analysis}, we only show the charts from RefineDet and YOLOv3 since RefineDet has similar patterns as other four detectors. RefineDet and other detectors show a clear maximum peak for TP count at confidence $>0.9$ and another weaker peak for confidence $<0.1$. While YOLOv3 has fewer predictions with high confidence. Therefore, although YOLOv3 is conservative predictions have high accuracy, it misses a large proportion of lesions and results in its low recall.

SSD yields the best adenomatous polyps detection recall, F1 score, and AP value. Overall, its mAP ($67.6\%$) ranks the third, closely matching the most recent detector ATSS and leading the following detector, Faster RCNN with mAP of $57.7\%$, by a considerable margin. For the harder task of hyperplastic polyp detection, RefineDet yields the highest scores for recall, F1, and AP. These results show that SSD-based detectors, SSD, RetinaNet, and RefineDet, are generally doing well in detecting polyps. RefineDet, by roughly adjusting anchors first, obtains better localization knowledge before generating final predictions. Faster RCNN has a similar two-step architecture. Therefore, it also has decent performance. This indicates the possibility to improve polyp detection performance by adding more refined location information before making final predictions. YOLOv4 outperforms YOLOv3 at almost every aspect, indicating that the tricks on general-purpose detectors are also effective on polyps detection. ATSS ranks second at hyperplastic polyp detection precision after YOLOv3. It also consistently performs on a par with RefineDet, especially in hyperplastic detection.

\subsection*{Generalizability and  Comparison with Previous Dataset}
The generalization ability refers to the adaptivity of the trained models to new, previously unseen data. This is very crucial in practical applications since the test images may have different distributions from the ones used to create the model. In order to test if the newly generated dataset can increase the generalizability of the trained models, we compare our results with the models only trained on a single dataset. 

We conduct the frame-based two-class polyp detection only on a single dataset, the KUMC dataset. The models are trained using the images from KUMC and tested on the full combined test set as in other experiments that consist of frames from different datasets. The results of different models are shown in Table~\ref{table:resControl}. We can see that, on average, the performance is dropped by $8\%$, when we compare the results in Table \ref{table:resFTPD} where all models are trained using the proposed dataset. The performance drop is mainly caused by the representativeness and the number of training samples. Although KUMC contains more variable sequences and frames than the other datasets combined, the color and illumination of different datasets may differ greatly, as shown in Figure \ref{plots:psample}. Therefore, the models trained on a single dataset may suffer poor generalization.

\begin{table}
\centering
\caption{\textbf{Result for training on KUMC and testing on the full combined test set}}
\begin{tabular}{ | c | r | r | r | r | r | r | r | r | }
    \hline
	             &\thead{Faster\\RCNN}&   SSD   &\thead{YOLO\\v3}&\thead{Retina\\Net}& DetNet  & RefineDet & YOLOv4 & ATSS  \\
    \hline
	mAP          &     \(52.7\)       &\(56.1\) &  \(42.6\)    &  \(36.9\)   &\(51.5\) & \(60.8\) &\(51.4\) & \(60.8\)  \\
    \hline

\end{tabular}
\label{table:resControl}
\end{table}

\subsection*{Frame-based One-class Polyp Detection} 
The results for detection only without classification are shown in Table~\ref{table:resFOPD}. We can see YOLOv3 achieves the highest precision among all detectors, which is consistent with the two-class results. With a reasonable recall, it also yields a high F1 score. Compared to its two-class detection performance, it is evidence that YOLOv3 is better at detecting than classifying polyps. YOLOv3 generates classification scores and bounding box adjustments at the same time. Since classification performance is based on the anchor information, YOLOv3's original anchors might not contain sufficient portions of a polyp due to its small size. We suggest that refined location information is more important for distinguishing polyp categories than for locating them.

\begin{table}
\centering
\caption{\textbf{Result for Frame-based One-class Polyp Detection}}
\begin{tabular}{ | c | r | r | r | r | r | r | }
    \hline
	             &  Precision &   Recall   & \thead{F1\\score}&    AP  & \thead{Inference\\Time}  & \thead{FPS}  \\
    \hline
	Faster RCNN  &  \(63.9\)  &  \(\mathbf{89.8}\)  &  \(74.7\)  &  \(85.6\) & \(52\)ms & \(19\)  \\
    \hline
	SSD          &  \(91.3\)  &  \(82.0\)  &  \(86.4\)  &  \(86.3\) & \textbf{\(\mathbf{17}\)ms} & \(\mathbf{59}\)  \\
    \hline
	YOLOv3       &  \(\mathbf{95.9}\)  &  \(78.0\)  &  \(86.0\)  &  \(81.0\) &  \textbf{\(\mathbf{17}\)ms} & \(\mathbf{59}\) \\
    \hline
	RetinaNet    &  \(86.1\)  &  \(86.6\)  &  \(86.3\)  &  \(87.9\) & \(61\)ms & \(16\)  \\
    \hline
	DetNet       &  \(85.8\)  &  \(81.8\)  &  \(83.7\)  &  \(80.5\) &   \(64\)ms & \(16\) \\
    \hline
	RefineDet    &  \(91.2\)  &  \(86.2\)  &  \(\mathbf{88.6}\)  &  \(\mathbf{88.5}\) & \(31\)ms & \(32\)  \\
    \hline
	YOLOv4    &  \(89.8\)  &  \(74.4\)  &  \(81.3\)  &  \(83.9\) & \(30\)ms & \(33\)  \\
    \hline
	ATSS    &  \(92.1\)  &  \(84.7\)  &  \(88.3\)  &  \(88.1\) & \(53\)ms & \(19\)  \\
    \hline

\end{tabular}
\label{table:resFOPD}
\end{table}

Table~\ref{table:resFOPD_detail} shows the detailed localization results for adenomatous and hyperplastic polyps. We can see that Faster RCNN achieves the best recall, which is the most important metric in clinical settings. For adenomatous polyps, Faster RCNN achieves $93.3\%$ recall, on a par with recent clinical screening results. It is one of the only three detectors (with RefineDet and ATSS) that achieve over $80\%$ recall for the hyperplastic polyps. Recall that in the above two-class detection, Faster RCNN also achieves the top three recall scores. Thanks to the region proposals, two-stage detectors usually have more chance to detect the polyps. While YOLOv3 also achieves competing performance in one-class detection. It yields the highest precision with reasonable recall score.

\begin{table}
\centering
\caption{\textbf{Frame-based One-class Polyp Detection Results for each Class}}
\begin{tabular}{ | c | c | r | r | r | r | }
    \hline
	                              & Category &  Precision &   Recall   &  F1-score &    AP   \\
    \hline \hline
	\multirow{2}{*}{Faster RCNN}  & ad &  \(75.9\)  &  \(\mathbf{93.3}\)  &  \(83.7\)  &  \(90.2\)  \\
    \cline{2-6}
                                  & hp &  \(50.0\)  &  \(\mathbf{84.3}\)  &  \(62.8\)  &  \(74.3\)  \\
    \hline \hline
	\multirow{2}{*}{SSD}          & ad &  \(96.1\)  &  \(87.6\)  &  \(91.7\)  &  \(89.8\)  \\
    \cline{2-6}
                                  & hp &  \(83.5\)  &  \(73.1\)  &  \(77.9\)  &  \(79.9\)  \\
    \hline \hline
	\multirow{2}{*}{YOLOv3}       & ad &  \(\mathbf{98.6}\)  &  \(86.4\)  &  \(92.1\)  &  \(\mathbf{90.5}\)  \\
    \cline{2-6}
                                  & hp &  \(\mathbf{90.5}\)  &  \(64.6\)  &  \(75.4\)  &  \(77.6\)  \\
    \hline \hline
	\multirow{2}{*}{RetinaNet}    & ad &  \(93.4\)  &  \(91.7\)  &  \(92.6\)  &  \(90.3\)  \\
    \cline{2-6}
                                  & hp &  \(75.2\)  &  \(78.5\)  &  \(76.8\)  &  \(81.4\)  \\
    \hline \hline
	\multirow{2}{*}{DetNet}       & ad &  \(93.5\)  &  \(86.6\)  &  \(89.9\)  &  \(81.7\)  \\
    \cline{2-6}
                                  & hp &  \(74.4\)  &  \(74.1\)  &  \(74.2\)  &  \(75.1\)  \\
    \hline \hline
	\multirow{2}{*}{RefineDet}    & ad &  \(96.1\)  &  \(89.6\)  &  \(\mathbf{92.7}\)  &  \(90.3\)  \\
    \cline{2-6}
                                  & hp &  \(83.7\)  &  \(80.8\)  &  \(82.2\)  &  \(\mathbf{85.2}\)  \\

    \hline \hline
	\multirow{2}{*}{YOLOv4}       & ad &  \(92.9\)  &  \(79.7\)  &  \(85.8\)  &  \(86.0\)  \\
    \cline{2-6}
                                  & hp &  \(84.4\)  &  \(65.9\)  &  \(74.0\)  &  \(79.3\)  \\
    \hline \hline
	\multirow{2}{*}{ATSS}    & ad &  \(96.1\)  &  \(87.3\)  &  \(91.5\)  &  \(89.6\)  \\
    \cline{2-6}
                                  & hp &  \(86.0\)  &  \(80.6\)  &  \(\mathbf{83.2}\)  &  \(84.2\)  \\
    \hline

\end{tabular}
\label{table:resFOPD_detail}
\end{table}

RefineDet still yields the best overall performance with the highest F1 score and AP. All SSD-based detectors perform almost equally well. The focal loss of RetinaNet does not show significant improvement on the original SSD model. DetNet does not show improvement over Faster RCNN, however, it makes the detector more balanced by increasing the precision by $20\%+$, resulting in a better F1 score.

We also evaluated the inference time of different models in frame-based one-class detection. All models are evaluated on an NVIDIA TESLA P100 GPU. As shown in Table~\ref{table:resFOPD}, The single-stage detectors (SSD, YOLOv3, RetinaNet, and RefineDet) are faster than the two-stage detectors (Faster RCNN and DetNet). SSD and YOLOv3 achieve the fastest inference time as 17ms, which is over 60 frames per second (fps). However, even for the slowest model DetNet, it still achieves 64ms, which is above 15 fps. Please note that a deeper backbone network will require more inference time than a shallower backbone network. For example, RetinaNet with ResNet-50 increases the inference time to 61ms from 17ms for SSD with VGG-16.

\begin{figure}
    \subfloat[RefineDet]{
    \begin{centering}
        \includegraphics[width=\columnwidth]{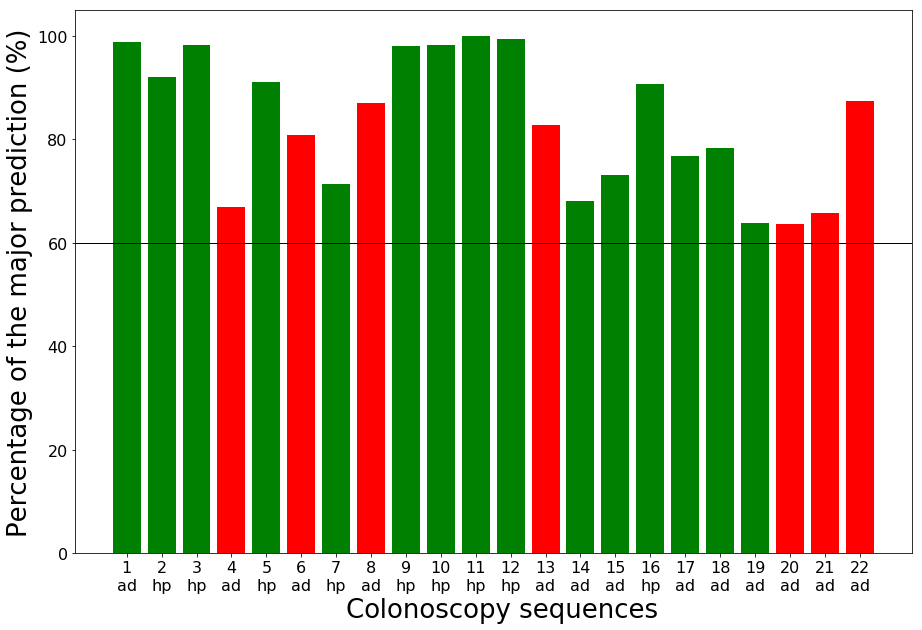}
        \label{plots:correct_percentage_refine}
    \end{centering}}
    \hfil
    
    \subfloat[RetinaNet]{
    \begin{centering}
        \includegraphics[width=\columnwidth]{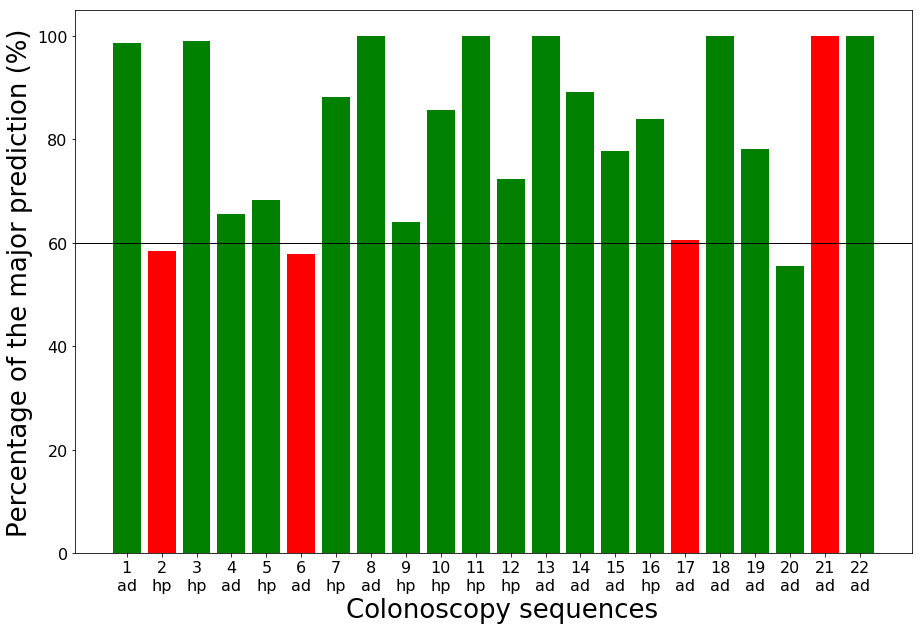} 
        \label{plots:correct_percentage_retina}
    \end{centering}}
    \hfil
    
	\caption{Percentage of the dominant class. Detectors predict the polyp category in each individual frame. The category with more than $50\%$ of all frames is the dominant category for that video sequence. The charts show the percentage of frames classified as the dominant class in each test sequence. (ad) and (hp) on the bottom means ground truth class adenomatous and hyperplastic respectively. Correct predictions are in green and misclassifications are in red. }
	\label{plot:correct_percentage}
\end{figure}

\subsection*{Sequence-based Two-class Polyp Classification} 
From Table~\ref{table:resSTPD}, we can see that both SSD, DetNet and YOLOv4 stand out in terms of precision, recall, and F1 score. This means that they are better at predicting correct polyp categories. Another interesting observation is that, although some detectors produce more consistent results for different frames in the same sequence, they do not yield higher precision. It becomes obvious when we plot the percentage of the dominant predicted category in each video sequence in Figure~\ref{plot:correct_percentage}. We only show the plots for RetinaNet and RefineDet as examples. DetNet, FasterRCNN, and RetinaNet are not very consistent in predicting the polyp class for some of the video sequences, with close to 50\% dominant class. This means the predictions are not robust with only a few frames to swing the result. RefineDet, SSD, YOLOv4, and ATSS, on the other hand, are relatively more robust in predicting the polyp class with most sequences above $70\%$.

\begin{table}
\centering
\caption{\textbf{Result for Sequence-based Two-class Polyp Classification}}
\begin{tabular}{ | c | r | r | r | r | }
    \hline
	             &  Precision &   Recall   &  F1-score   \\
    \hline
	Faster RCNN  &  \(81.2\)  &  \(81.2\)  &  \(81.2\)  \\
    \hline
	SSD          &  \(86.6\)  &  \(85.0\)  &  \(85.8\)  \\
    \hline
	YOLOv3       &  \(72.2\)  &  \(60.3\)  &  \(65.7\)  \\
    \hline
	RetinaNet    &  \(81.8\)  &  \(82.9\)  &  \(82.4\)  \\
    \hline
	DetNet       &  \(85.8\)  &  \(\mathbf{86.8}\)  &  \(\mathbf{86.3}\)    \\
    \hline
	RefineDet    &  \(78.1\)  &  \(73.1\)  &  \(75.5\) \\
    \hline
	YOLOv4    &  \(\mathbf{87.5}\)  &  \(80.8\)  &  \(84.0\) \\
    \hline
	ATSS    &  \(81.2\)  &  \(81.2\)  &  \(81.2\) \\
    \hline

\end{tabular}
\label{table:resSTPD}
\end{table}

\section*{Conclusion}
In this paper, we have developed a relatively large endoscopic dataset for polyp detection and classification. We have also evaluated and compared the performance of eight state-of-the-art deep learning-based object detectors. Our results show that deep CNN models are promising in assisting CRC screening. Without too much modification, general object detectors have already achieving adenomatous polyps detection sensitivity of $91\%$ in the one-class detector and around $70\%$ precision in the classification task. Among all the detectors we have tested, YOLOv4, ATSS, and RefineDet perform relatively well in all tests with balanced precision and recall scores and consistent results for the same lesions. Our experiments also show the refinement of location information before classification will effectively boost the performance. 

This study can serve as a baseline for future research in polyp detection and classification. The developed dataset can serve as a standardized platform and help researchers to design more specialized CNN models for polyp classification.  Looking back at the fast development in the computer vision field in recent years, the availability of the benchmark dataset plays a significant role. We hope our dataset will greatly facilitate the computer-aided diagnosis of colorectal cancer.



%
%
%


\end{document}